\documentclass[11pt]{article}
\usepackage{fullpage}

\usepackage[utf8]{inputenc} % allow utf-8 input
\usepackage[T1]{fontenc}    % use 8-bit T1 fonts
\usepackage{hyperref}       % hyperlinks
\usepackage{url}            % simple URL typesetting
\usepackage{booktabs}       % professional-quality tables
\usepackage{amsfonts}       % blackboard math symbols
\usepackage{nicefrac}       % compact symbols for 1/2, etc.
\usepackage{microtype}      % microtypography
\usepackage{tcolorbox}
\usepackage{diagbox}

\usepackage{enumerate}
\usepackage[shortlabels]{enumitem}
\usepackage{algorithmic}
\usepackage{algorithm}
\usepackage{subfigure}
\usepackage{graphicx} % more modern
\usepackage{caption}
\usepackage{amsmath}
\usepackage{amsthm}
\usepackage{amssymb}
\usepackage{tikz}
\usepackage{tablefootnote}
\usepackage{multirow}
\usepackage{enumerate}
\usepackage{color}
\usepackage{xcolor}
\usepackage{natbib}

\usetikzlibrary{arrows}

\allowdisplaybreaks[4]

\usepackage{mathrsfs}

\usepackage{hyperref}
\usepackage{bm,todonotes}

\allowdisplaybreaks

\hypersetup{
	colorlinks=true,
	filecolor=blue,
	citecolor = blue,
	urlcolor=cyan,
}

\usepackage{amsmath,amsfonts,bm}

\def\1{\bm{1}}

\DeclareMathAlphabet{\mathsfit}{\encodingdefault}{\sfdefault}{m}{sl}
\SetMathAlphabet{\mathsfit}{bold}{\encodingdefault}{\sfdefault}{bx}{n}

\def\Xi{\boldsymbol{\xi}}

\def\Phi{\boldsymbol{\phi}}

\def\0{{\bf 0}}
\def\1{{\bf 1}}

\title{
Learning Personalized Brain Functional Connectivity of MDD Patients from Multiple Sites via Federated Bayesian Networks
}

\author{
Shuai Liu \thanks{Department of Information Systems and Intelligent Business, School of Management, Xi'an Jiaotong University; email: \texttt{hljliushuai@stu.xjtu.edu.cn}. } \\
\and
Xiao Guo \thanks{Corresponding author: Center for Modern Statistics, School of Mathematics, Northwest University; email: \texttt{xiaoguo@nwu.edu.cn}. }\\
 \and
Shun Qi \thanks{Key Laboratory of Biomedical Information Engineering of Ministry of Education, Institute of Health and Rehabilitation Science, School of Life Science and Technology, Xi'an Jiaotong University; email: \texttt{qishun@xjtu.edu.cn}. }\\
	\and
Huaning Wang  \thanks{Department of Psychiatry, Xijing Hospital, Air Force Medical University; email: \texttt{xskzhu@fmmu.edu.cn}. }\\
	\and
  Xiangyu Chang \thanks{Center for Intelligent Decision-Making and Machine Learning, School of Management, Xi’an Jiaotong University; email: \texttt{xiangyuchang@xjtu.edu.cn}. } 
}

\begin{document}

\maketitle

\begin{abstract}
Identifying functional connectivity biomarkers of major depressive disorder (MDD) patients is essential to advance understanding of the disorder mechanisms and early intervention.
However, due to the small sample size and the high dimension of available neuroimaging data, the performance of existing methods is often limited. 
Multi-site data could enhance the statistical power and sample size, while they are often subject to inter-site heterogeneity and data-sharing policies.
In this paper, we propose a federated joint estimator, NOTEARS-PFL, for simultaneous learning of multiple Bayesian networks (BNs) with continuous optimization, to identify disease-induced alterations in MDD patients.
We incorporate information shared between sites and site-specific information into the proposed federated learning framework to learn personalized BN structures by introducing the group fused lasso penalty. 
We develop the alternating direction method of multipliers, where in the \emph{local update} step, the neuroimaging data is processed at each local site. Then the learned network structures are transmitted to the center for the \emph{global update}. 
In particular, we derive a closed-form expression for the local update step and use the iterative proximal projection method to deal with the group fused lasso penalty in the global update step.
We evaluate the performance of the proposed method on both synthetic and real-world multi-site rs-fMRI datasets.  
The results suggest that the proposed NOTEARS-PFL yields superior effectiveness and accuracy than the comparable methods.
\end{abstract}

\section{Introduction}
\label{sec:introduction}
Major depressive disorder (MDD) is one of the most prevalent, costly, and disabling mental disorders worldwide \citep{cassano2002depression,jia2010high}. 
It is often characterized by persistent sadness, worthlessness, hopelessness, anxiety, and cognitive impairments \citep{fu2008pattern}.
The negative psychology of MDD can lead to severe consequences, including suicide \citep{jia2015impact}.
The productivity loss and disability due to MDD also pose a severe burden to society and affected families \citep{vos2017global}.
Therefore, understanding the pathogenesis of MDD is crucial for effective intervention, diagnosis, and treatment \citep{belmaker2008major}. 
Diagnostic neuroimaging  has been shown to be an effective modality to understand the underlying pathological mechanisms and cognitive information of brain diseases \citep{liu2016relationship,liu2017view}, and has been used for early identification of MDD \citep{smith2013functional}.
Resting-state functional magnetic resonance imaging (rs-fMRI) is one of the most widely used imaging modes for MDD identification \citep{hamilton2011investigating, liu2013abnormal,lin2016cca}.
It can detect abnormal brain functional connectivity in MDD patients, allowing non-invasive investigation of the disease \citep{ye2015changes}.
Functional connectivity (FC), an essential biomarker in neurological disorders, is traditionally defined as the Pearson correlation between different regions of interest (ROIs) \citep{ryali2012estimation}.
However, this association still does not truly determine the direction of interactions between ROIs. An accurate understanding of this directional information is critical to understanding the brain functional integration of MDD \citep{price2017data}.
To further investigate FC with rs-fMRI data, a number of approaches beyond the Pearson correlation are proposed.
For example, \cite{kandilarova2018altered} used a dynamic causal modeling (DCM) method to reveal differences of FC in anterior insula between healthy control (HC) participants and MDD patients.
\cite{hamilton2011investigating} used Granger causality (GC) analysis  method to investigate the aberrant patterns of FC in MDD, and identify the limbic inhibition of dorsal cortex as a key biomarker for MDD diagnosis .
Nevertheless, these methods are not robust for analyzing specific individuals or large-scale ROIs \citep{molenaar2004manifesto,friston2011network}.

Bayesian network (BN), an approach for studying the conditional dependencies between variables in a system via a directed acyclic graph (DAG), has been used for detecting abnormal brain FC in MDD patients \citep{koller2009probabilistic,ide2014bayesian,liu2022learning}.
Compared with DCM and GC methods, BN benefits from utilizing more samples to improve the robustness and stability of the resulting FC. 
Note that the amount of rs-fMRI data collected at a single imaging site is limited due to the difficult and expensive data acquisition in practice.
Therefore, it is difficult to provide enough data from a single site to make a good BN estimate.
Fortunately, multiple rs-fMRI datasets from different sites provide the possibility for enhancing the quality of learned BNs. 
However, multi-site data are often subject to inter-site heterogeneity and data-sharing policies \citep{li2020multi,tong2022distributed}.
How to learn more helpful heterogeneous  information from multi-site data with considering data-sharing policies is still one of the research hotspots of BN learning.

In this paper, we consider the problem of learning BN structures using multiple datasets.
Our initial interest arises from discovering noninvasive biomarkers inherent in multi-site rs-fMRI data.
Generally, data collected at multiple imaging sites have a common disease of interest (such as MDD), which results in that they may share similar disease-induced connectivity abnormalities. 
Therefore, we consider leveraging data from multi-site to learn multiple BNs.
We are committed to addressing the three challenges: (1) {\textbf{computational inefficiency} of the structural learning algorithms of BNs, which partially comes from the DAG constraint and the resulting  combinatorial optimization problem;}
(2) \textbf{large heterogeneity} of multi-site rs-fMRI data, which is caused by inter-site variation in scanners, image acquisition protocols, and patient populations across sites \citep{wang2022multi}; 
(3) \textbf{data sharing barriers}, that is, patients worry that sharing their medical data will lead to personal information disclosure \citep{li2020multi}. The government is likely to promulgate some privacy protection regulations and inflict severe penalties for medical data breach events \citep{jin2019review}.

\begin{figure}[h]
     \centering
  {\includegraphics[width=1\textwidth]{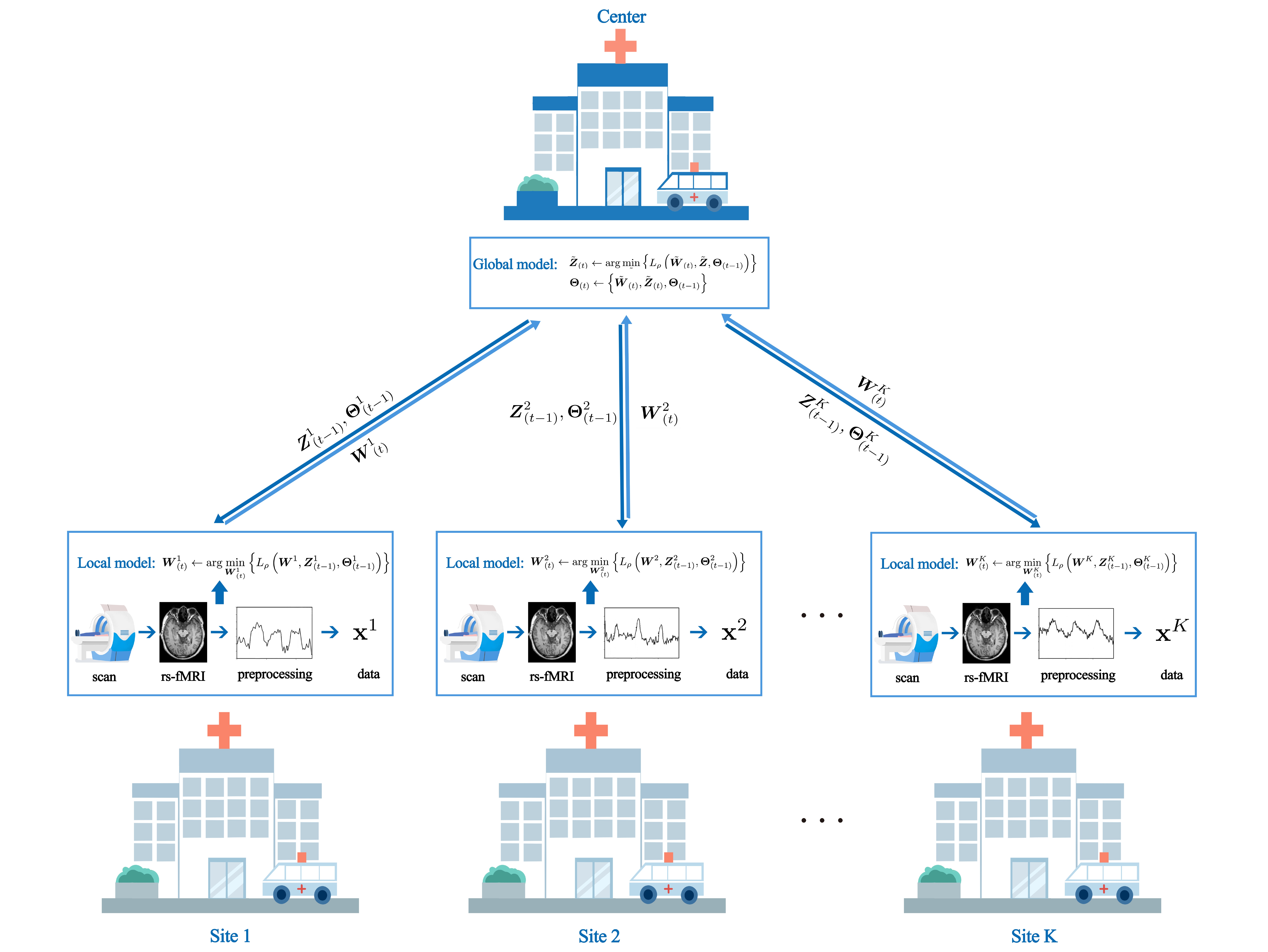}}
  \caption{Illustration of the learning framework of NOTEARS-PFL. Each site stores and processes rs-fMRI data locally. In the $t$-th iteration, each site estimates the local weighted adjacency matrix $\boldsymbol{W}^{k}_{(t)}$ through the local update step. Then, each site transmits the estimated local matrix to the center. Center updates the global weighted adjacency matrices $\tilde{\boldsymbol{Z}}_{(t)}$  through the global update step based on the received $\tilde{\boldsymbol{W}}_{(t)}=\{\boldsymbol{W}^{1}_{(t)}, \ldots, \boldsymbol{W}^{K}_{(t)}\}$. Finally, the center transmits the estimated global matrix to each site for the $(t+1)$-th iteration update.}
  \label{fig: framework}
\end{figure}

To address the above issues, we propose a federated joint estimator for the simultaneous learning of multiple BNs with continuous optimization.
In pursuit of computational efficiency, we convert the traditional combinatorial BN structure learning problem into a continuous numerical optimization problem inspired by the NOTEARS method \citep{zheng2018dags}.
To capture the heterogeneity, we learn multiple BNs from multi-site data but with similar structures by introducing the group fused lasso penalty instead of learning a single BN.
To lighten the data-sharing barriers, we study the structure learning of multiple BNs in the federated learning (FL) setting, which is a natural method to tackle the data-sharing problem  \citep{pfitzner2021federated}. 
We refer to the proposed \textbf{NOTEARS}-based \textbf{P}ersonalized \textbf{F}ederated \textbf{L}earning framework of learning multiple BNs' structures as NOTEARS-PFL, whose whole protocol is shown in Figure \ref{fig: framework}.

\newpage

The merits of the proposed NOTEARS-PFL are multi-fold.

\begin{itemize}
    \item NOTEARS-PFL is capable of learning site-specific BNs from multi-site rs-fMRI data, which makes full use of the multi-site data while still being flexible enough to capture the site-specific BN structures. 

    \item NOTEARS-PFL is subtly fitted in an FL framework without sharing and exchanging the original data.  
    
    \item NOTEARS-PFL overcomes the computational intractability of traditional BNs by transforming the discrete combinatorial optimization to continuous numerical optimization. 
    
    \item NOTEARS-PFL can effectively detect the common abnormal FC in MDD patients among multiple sites, especially the connectivity between the insula, thalamus cortex, and middle temporal gyrus. 
    Meanwhile, NOTEARS-PFL can also identify the site-specific abnormal FC due to the differences in scanners, protocols, and patient populations.
    
\end{itemize}

The paper is organized as follows. In Section \ref{sec:related_works}, we briefly review directed FC estimation methods for rs-fMRI data and BN structure learning with continuous optimization. 
We present the formulation and optimization approach of NOTEARS-PFL in Section \ref{sec:formulation}. 
We conduct simulation experiments to validate NOTEARS-PFL in Section \ref{sec:experiments}, and analyze the real-world multi-site rs-fMRI data in Section \ref{sec: real_world}.
Finally, we conclude this paper in Section \ref{sec:conclusion}.

\section{Related Work}\label{sec:related_works}

\subsection{Directed FC Estimation Methods for rs-fMRI Data}
The directed FC, also known as effective FC, helps depict the abnormalities or dysfunction of the brain connectivity network \citep{smith2012future} and shows how information flows in the brain connectivity network \citep{henry2017causal}.
Based on rs-fMRI data, many directed FC estimation methods have been developed, including the following four types.

(1) \textit{ Granger causality (GC)}. 
GC establishes a model in the framework of vector autoregression (VAR) and determines the edges to be added after considering the autoregressive effect of each region \citep{granger1969investigating,liao2009kernel}. 
GC is applied to detect the effects of psychostimulants on youths with attention deficit hyperactivity disorder (ADHD) \citep{peterson2009fmri}.
However, when GC is applied to analyze specific individuals, heterogeneity across individuals may lead to spurious results \citep{molenaar2004manifesto}.

(2) \textit{Dynamic Causal Modeling (DCM)}. 
DCM is a method to capture nonlinear relationships between brain regions that may be affected by different stimuli during scanning \citep{friston2003dynamic}, and has been used to detect the propagation of epileptic seizures \citep{murta2012dynamic}. 
The limitation of DCM  is that it cannot handle large amounts of ROIs or analysis group rs-fMRI data. 
 \citep{friston2011network}.

(3) \textit{Constraint-based methods}. 
Constraint-based methods estimate directed edges by conditional independence tests.
Common constraint-based methods include Peter Spirtes, and Clark Glymour (PC) algorithm \citep{spirtes1991algorithm}, conservative PC (CPC) \citep{ramsey2006adjacency}, fast causal inference (FCI) \citep{spirtes2001anytime}, and cyclic causal discovery (CCD) \citep{richardson1996polynomial}.
The PC algorithm can obtain reliable networks on data aggregated across individuals \citep{smith2011network}.
However, according to results from \cite{smith2011network}, PC can detect the existence of edges but can not accurately determine the directions at the individual level.

(4) \textit{Score-based methods}. 
Score-based methods define a score function that measures how well the BN structure fits the observed data and searches for the best network structure.
Standard score-based methods include greedy equivalent search (GES) \citep{chickering2002optimal}, independent multiple-sample greedy equivalent search (IMaGES) \citep{ramsey2010six}, and group iterative multiple model Estimation (GIMME) \citep{gates2012group}.
Score-based methods typically require large samples to identify edges and directions, which limits the utility of these methods when using rs-fMRI data for individual-level analysis \citep{mumford2014bayesian}.
Meanwhile, the performance of score-based methods is poor on data with a large number of ROIs, which makes it difficult to analyze the complete parcellation of the brain.
Due to the combinatorial acyclicity constraint of DAG, it is often hard to find BN with the optimal score. And its scalability is also limited \citep{liu2022learning}.

\subsection{BN Structure Learning with Continuous Optimization}

In this work, we focus on BN learning. BN is a probabilistic graphical model that captures dependencies among a collection of random variables.
Each model is associated with a directed network $\mathcal{G}=(V,E)$, where the vertex set $V$ corresponds to the random variables and the edge set $E$ corresponds to the set of conditional independence \citep{drton2017structure}. 

BN models the conditional independence among random variables via \emph{Markov properties} \citep{cowell1998introduction,marrelec2004estimation}.
Denote $\boldsymbol{X}=(X_{1}, \ldots, X_{d})^\top$ that satisfies the \emph{local Markov property} with respect to a DAG $\mathcal G$ if 
$$X_v\bot X_{\mbox{\scriptsize nd}_\mathcal G(v)\backslash \mbox{\scriptsize pa}_\mathcal G(v) }\mid X_{\mbox{\scriptsize pa}_\mathcal G(v) },$$
where ${\rm pa}_\mathcal G(v)=\{w\in V:\{w,v\}\in E\}$ denotes the parents of node $v$ in $\mathcal G$, ${\rm nd}_\mathcal G(v)=V\backslash \mbox{de}_\mathcal G(v) $ denotes the non-descendants of $v$ in $\mathcal G$ with $\mbox{de}_\mathcal G(v)=\{w\in V:w=v\; \mbox{or}\; v\rightarrow \ldots \rightarrow w\;\mbox{in}\; G\}$ representing the descendants of $v$ in $\mathcal G$.

Alternatively, if $\boldsymbol{X}$ is a continuous random vector and it has a density $\mathbb P(\boldsymbol X)$ with respect to a product measure, then the local Markov property is equivalent to the following \emph{factorization property} \citep{verma1990causal},
\begin{equation}
\mathbb{P}(\boldsymbol{X})=\prod_{v=1}^{d} \mathbb{P}\left(X_{v} \mid X_{\mbox{\scriptsize pa}_\mathcal G(v)}\right).
\label{eq:Pxjoint}
\end{equation}

The structure learning of BN is to estimate the DAG $\mathcal G$ using the observations from the density $\mathbb P(\boldsymbol X)$. 
Score-based methods seek the optimal $\mathcal{G}$ by minimizing the score function $\mathcal{S}(W)$ with respect to the corresponding weighted adjacency matrix $\boldsymbol{W}\in \mathbb R^{d\times d}$, subject to that the induced graph $g(\boldsymbol{W})$ is a DAG; see the LHS of equation \eqref{eq:convert}. The DAG constraint initiates a combinatorial optimization problem, which is computationally costly when $d$ is large.  

Recently, \cite{zheng2018dags} proposed a NOTEARS method which enforces the acyclicity via setting a smooth function $h(\boldsymbol{W})$ exactly at zero; see the RHS of equation \eqref{eq:convert}. By this novel characterization of acyclicity, the resulting optimization problem can be solved efficiently using standard algorithms.

\begin{equation}
    \begin{array}{cl}
             \min _{\boldsymbol{W}\in \mathbb R^{d\times d}} \mathcal{S}(\boldsymbol{W}), & \min _{\boldsymbol{W}\in \mathbb R^{d\times d}} \mathcal{S}(\boldsymbol{W}), \\
     \text { s.t. } g(\boldsymbol{W}) \in \mathrm{DAGs}, & \text { s.t. } h(\boldsymbol{W})=0.
   \end{array}
   \label{eq:convert}
\end{equation}

Currently, NOTEARS has been extended to handle problems in many situations.
For example, DAG Graph Neural Network (DAG-GNN) extends NOTEARS to solve nonlinear cases by incorporating neural networks \citep{yu2019dag}.
Graph AutoEncoder (GAE) extends NOTEARS and DAG-GNN into a graph autoencoder model, facilitating vector-valued variables \citep{ng2019graph}.
\cite{pamfil2020dynotears} proposed DYNOTEARS method to learn dynamic BNs from time-series data.
 \cite{huang2020causal} used reinforcement learning to find the optimal DAGs from incomplete data based on NOTEARS.

To the best of our knowledge, few studies have focused on using NOTEARS to learn personalized BNs from fMRI data.
\cite{zhang2022detecting} proposed a joint BN estimation model to  detect abnormal directed FC in schizophrenia (SZ) patients.
However, the joint estimation model proposed by \cite{zhang2022detecting} is not applicable for the FL setting. 
 \cite{ng2022towards} proposed NOTEARS-ADMM model to estimate the structure of BN in FL setting. However, they did not consider the data heterogeneity, which can lead to model bias \citep{zhao2018federated,jiang2022harmofl}.
Compared with \cite{zhang2022detecting} and \cite{ng2022towards}, the proposed NOTEARS-PFL is powerful because it captures the data heterogeneity, overcomes the data sharing barriers, and attains computational efficiency, simultaneously.

\section{NOTEARS-based Personalized Federated Learning Framework}\label{sec:formulation}
\subsection{Preliminary}

The BN associated with a DAG $\mathcal G$ can also be thought of as a structural equation model~\citep{bollen1989structural}
\begin{equation}
X_{i}=f_i(W_{i}^{T} \boldsymbol{X})+Z_{i}, i=1,2, \ldots, d,
\label{eq:ANMs}
\end{equation}
where $\boldsymbol{W}=\left(W_{1},\ldots, W_{d}\right)\in\mathbb{R}^{d \times d}$ denotes the weighted adjacency matrix whose nonzero coefficients correspond to the directed edges in $\mathcal{G}$,  $f_i$'s are functions related to the conditional distributions of $X_{i}$'s, and $Z_{i}$’s are jointly independent
random noises. 
Equation \eqref{eq:ANMs} is equivalent to the local Markov property and factorization property introduced in Section \ref{sec:related_works} \citep{drton2017structure}.  

We consider the linear structural equation model 
\begin{equation}
X_{i}=W_{i}^{T} \boldsymbol{X}+Z_{i}, i=1,2, \ldots, d.
\label{eq:Xi}
\end{equation}
Actually, the multivariate Gaussian distribution can be modeled as equation \eqref{eq:Xi} with $Z_i$'s being independent Gaussian noises.

Let the data matrix $\mathbf{x}\in\mathbb{R}^{n \times d}$.
\cite{zheng2018dags} formulates the structure learning of BNs as 

\begin{equation}
\begin{aligned}
\min _{\boldsymbol{W}\in\mathbb{R}^{d \times d}} \quad &\frac{1}{2 n} \left\|\mathbf{x}-\mathbf{x}\boldsymbol{W} \right\|_{F}^{2}+\lambda\|\boldsymbol{W}\|_{1},\\
\mbox{s.t.}\quad  &h(\boldsymbol{W})=0,
\end{aligned}
\label{eq:notears}
\end{equation}
where
\begin{equation}
h(\boldsymbol{W}):= \operatorname{tr}\left(e^{\boldsymbol{W} \odot \boldsymbol{W}}\right)-d=0,
\label{eq:constraint}
\end{equation}
$\odot$ denotes the Hadamard product, $e^{A}=\sum_{k=0}^{\infty} \frac{1}{k !} A^k$ denotes the the matrix exponential operator \citep{leonard1996matrix}, $\|A\|_{F}$ denotes the Frobenius norm, $\|A\|_{1}$ refers to the entry-wise $\ell_{1}$ norm  and the tuning parameter $\lambda>0$ controls the amount of sparsity.

\subsection{Problem Formulation}

We now proceed to formulate our NOTEARS-PFL method, which can efficiently learn multiple heterogeneous BNs without sharing data. 

We consider the setting in which there is a fixed set of $K$ sites in total, and each site has its own local dataset. 
The $k$-th site holds $n_{k}$ i.i.d. samples, denoted by $\mathbf{x}^{k}=\left\{x^{k }_{i}\right\}_{i=1}^{n_{k}}\in\mathbb{R}^{{n_k}\times d}$. Let $\mathcal{D}=\{\mathbf{x}^{1},\dots,\mathbf{x}^{K}\}$ be the $K$ overall observational datasets.
The goal is to infer the $K$ weighted adjacency
matrices $\tilde{\boldsymbol{W}}_{0}=\{\boldsymbol{W}^{1}_{0}, \ldots, \boldsymbol{W}^{K}_{0}\}$ (correspond to the BN structures)
using $\mathcal{D}$.

To incorporate the heterogeneity among $K$ sites, we consider the following optimization problem:
\begin{equation}
\begin{aligned}
\min _{\tilde{\boldsymbol{W}}} \quad &\sum_{k=1}^{K}  \frac{1}{2 n_{k}} \left\|\mathbf{x}^{k}-\mathbf{x}^{k}\boldsymbol{W}^{k} \right\|_{F}^{2}+\mathcal{R}(\boldsymbol{\tilde{W}}),\\
\quad\quad\quad\mbox{s.t.}\quad  &h({\boldsymbol W}^{k})=0,\ k=1,\dots, K,
\end{aligned}
\label{eq:mdag}
\end{equation}
where $h({\boldsymbol W}^{k})$ is defined in the same way as equation \eqref{eq:constraint} and $\mathcal{R}(\tilde{\boldsymbol{W}})$ is a penalty function for inducing the similarity and sparsity within $\tilde{\boldsymbol{W}}$. 

With the data sharing constraint, we formulate equation \eqref{eq:mdag} into the framework of ADMM, which is a natural solution to distributed learning and FL \citep{ng2022towards}.
Define a set of consensus variables $\tilde{\boldsymbol{Z}}=\{{\boldsymbol{Z}^{1}, \ldots, \boldsymbol{Z}^{K}}\}$, which can be thought of as the \emph{global} weighted adjacency matrices, in contrast to the \emph{local} weighted adjacency matrices $\tilde{\boldsymbol{W}}$. 
Then we rewrite equation \eqref{eq:mdag} as
\begin{equation}
\begin{aligned}
\min _{\tilde{\boldsymbol{W}},\tilde{\boldsymbol{Z}}} \quad &\sum_{k=1}^{K}  \mathcal{L}(\boldsymbol{W}^{k},\mathbf{x}^{k})+\mathcal{R}(\tilde{\boldsymbol{Z}}),\\
\mbox{s.t. }\quad  &h(\boldsymbol{Z}^{k})=0,\\
&\boldsymbol{W}^{k}=\boldsymbol{Z}^{k},\quad k=1, \ldots, K,
\end{aligned}
\label{eq:fldag}
\end{equation}
where
\begin{equation}
\mathcal{L}(\boldsymbol{W}^{k},\mathbf{x}^{k}):= \frac{1}{2 n_{k}} \left\|\mathbf{x}^{k}-\mathbf{x}^{k}\boldsymbol{W}^{k} \right\|_{F}^{2},
\label{eq:fldag-loss}
\end{equation}
$h({\boldsymbol Z}^{k})$ is defined in the same way as equation \eqref{eq:constraint}, and $\mathcal{R}(\tilde{\boldsymbol{Z}})$ is defined by

\begin{equation}
\mathcal{R}(\tilde{\boldsymbol{Z}}):=\lambda_{1}\sum_{k=1}^{K}\|\boldsymbol{Z}^{k}\|_{1}+\lambda_{2} \sum_{k=1}^{K-1} \|\boldsymbol{Z}^{k+1}-\boldsymbol{Z}^{k}\|_{F},
\label{eq:fused}
\end{equation}
where $\lambda_{1}>0$ controls the sparsity of $\boldsymbol{Z}^{k}$ and $\lambda_{2}>0$ controls the similarity within $\tilde{\boldsymbol{Z}}$.

The general optimization procedure can be summarized as follows. First, each site estimates the local weighted adjacency matrix $\boldsymbol{W}^{k}$ by using local dataset $\mathbf{x}^{k}$.
Then, instead of transmitting the original data, each site only transmits its local weighted adjacency matrix to the center.
After that, the center updates the global weighted adjacency matrices $\tilde{\boldsymbol{Z}}_{0}=\{{\boldsymbol{Z}^{1}_{0}, \ldots, \boldsymbol{Z}^{K}_{0}}\}$, based on received $\tilde{\boldsymbol{W}}_{0}$,
Finally, the center transmits the updated $\tilde{\boldsymbol{Z}}$ to each site, which performs the next update locally based $\boldsymbol{W}^{k}=\boldsymbol{Z}^{k}$.

In the next subsection, we provide the computational details of NOTEARS-PFL.

\subsection{Optimization Algorithm}
By applying the augmented Lagrangian method \citep{bertsekas2014constrained}, the equation \eqref{eq:fldag} can be written as

\begin{equation}
\begin{aligned}
L\left(\tilde{\boldsymbol{W}},\tilde{\boldsymbol{Z}},\boldsymbol{\Theta}\right)=&\sum_{k=1}^{K}  \mathcal{L}(\boldsymbol{W}^{k},\mathbf{x}^{k})+\mathcal{R}(\tilde{\boldsymbol{Z}})+\sum_{k=1}^{K}\alpha_{k} h(\boldsymbol{Z}^{k})+\frac{\rho_{1}}{2}\sum_{k=1}^{K}|h(\boldsymbol{Z}^{k})|^{2}+\sum_{k=1}^{K} \operatorname{tr}(\beta_{k}(\boldsymbol{W}^{k}-\boldsymbol{Z}^{k})^{\top})\\
&+\frac{\rho_{2}}{2} \sum_{k=1}^{K}\left\|\boldsymbol{W}^{k}-\boldsymbol{Z}^{k}\right\|_{F}^{2},
\end{aligned}
\label{eq:fldag_al}   
\end{equation} 
where $\tilde{\boldsymbol{\Theta}}=\{\rho_{1},\rho_{2}, \alpha_{1}, \ldots, \alpha_{K}, \beta_{1}, \ldots, \beta_{K}\}$, $\rho_{1},\rho_{2}$ are the penalty coefficients, $\alpha_{1}, \ldots, \alpha_{K}$ and $\beta_{1}, \ldots, \beta_{K}$ are the Lagrange multiplies. 

At the $t$-th iteration, we can solve equation \eqref{eq:fldag_al} as follows:

 \textbf{(1) Local update}:
\begin{equation}
\tilde{\boldsymbol{W}}_{(t)} \leftarrow \arg \min_{\tilde{\boldsymbol{W}}}\left\{L\left(\tilde{\boldsymbol{W}},\tilde{\boldsymbol{Z}}_{(t-1)},\tilde{\boldsymbol{\Theta}}_{(t-1)}\right)\right\}.
\label{eq:W_t}
\end{equation}

\textbf{(2) Global update}:
\begin{equation}
\tilde{\boldsymbol{Z}}_{(t)} \leftarrow \arg \min_{\tilde{\boldsymbol{Z}}} \left\{L\left(\tilde{\boldsymbol{W}}_{(t)},\tilde{\boldsymbol{Z}},\tilde{\boldsymbol{\Theta}}_{(t-1)}\right)\right\}.
\label{eq:Z_t}
\end{equation}

 \textbf{(3) Global update}:
\begin{equation}
\tilde{\boldsymbol{\Theta}}_{(t)} \leftarrow \left\{\tilde{\boldsymbol{W}}_{(t)},\tilde{\boldsymbol{Z}}_{(t)},\tilde{\boldsymbol{\Theta}}_{(t-1)}\right\}.
\label{eq:Theta}
\end{equation}
\
\subsubsection{Federated Local Update of $\tilde{\boldsymbol{W}}$}

Taking the derivative of equation \eqref{eq:fldag_al} with respect to $\tilde{\boldsymbol{W}}$, we can update each $\boldsymbol{W}^{k}_{(t)}$ of $\tilde{\boldsymbol{W}}_{(t)}$ as   

\begin{equation}
{\boldsymbol{W}^{k}_{(t)}}:=\underset{\boldsymbol{W}^{k}}{\arg \min }( \mathcal{L}(\boldsymbol{W}^{k},\mathbf{x}^{k})+\frac{\rho_{2,(t-1)}}{2}\left\|\boldsymbol{W}^{k}-\boldsymbol{Z}^{k}_{(t-1)}\right\|_{F}^{2}+\operatorname{tr}(\beta_{k,(t-1)}(\boldsymbol{W}^{k}-\boldsymbol{Z}^{k}_{(t-1)})^{\top})).
\label{eq:W_k}
\end{equation}   
Let $U^{k}=\frac{1}{n_{k}}(\mathbf{x}^{k})^{\top} \mathbf{x}^{k}$ and assume $U^{k}+\rho_{2} I$ is invertible.
Equation \eqref{eq:W_k} has a closed form expression

\begin{equation}
\boldsymbol{W}^{k}_{(t)}=(U^{k}+\rho_{2,(t-1)} I)^{-1}(\rho_{2,(t-1)} \boldsymbol{Z}^{k}_{(t-1)}-\beta_{k,(t-1)}+U^{k}),
\label{eq:W_k_hat}
\end{equation}
where the details of the derivation are given in Appendix \ref{sec:local_update}.

\subsubsection{Federated Global Update of $\tilde{\boldsymbol{Z}}$}
Similar to equation \eqref{eq:W_k}, we can update each $\boldsymbol{Z}^{k}_{(t)}$ of  $\tilde{\boldsymbol{Z}}_{(t)}$ as  
\begin{equation}
\boldsymbol{Z}^{k}_{(t)}:=\underset{\boldsymbol{Z}^{k}}{\arg \min }\{\mathcal{L}(\tilde{\boldsymbol{Z}})+\mathcal{R}(\tilde{\boldsymbol{Z}})\},
\label{eq:Z_k_hat}
\end{equation}
where
\[
\begin{aligned}
\mathcal{L}(\tilde{\boldsymbol{Z}}):=&\sum_{k=1}^{K}\alpha_{k,(t-1)} h(\boldsymbol{Z}^{k})+\frac{\rho_{1,(t-1)}}{2}\sum_{k=1}^{K}|h(\boldsymbol{Z}^{k})|^{2}+\frac{\rho_{2,(t-1)}}{2}\sum_{k=1}^{K}\left\|\boldsymbol{W}^{k}_{(t)}-\boldsymbol{Z}^{k}\right\|_{F}^{2})\\
&+\sum_{k=1}^{K}\operatorname{tr}(\beta_{k,(t-1)}(\boldsymbol{W}^{k}_{(t)}-\boldsymbol{Z}^{k})^{\top}).
\end{aligned}
\label{eq: gradien}
\]   

\begin{figure}[]
		\renewcommand{\algorithmicrequire}{\textbf{Input:}}
		\renewcommand{\algorithmicensure}{\textbf{Output:}}
		\begin{algorithm}[H]
			\caption{Dykstra-like iterative proximal algorithm (DIPA)}
			\begin{algorithmic}[1]\label{algorithm: DIPA}
				\REQUIRE $K$: Number of site;  $\tilde{\boldsymbol{Z}}$:  global adjacency matrices; $\tilde{\boldsymbol{W}}$:  local adjacency matrices; $N$: The number of iterations; $\epsilon$: Tolerance.
					\ENSURE
		    $\tilde{\boldsymbol{Z}}$: The result of equation  \eqref{eq: proximal_operator}.
		       \STATE $\nabla \mathcal{L}_{k}\left(\boldsymbol{Z}^{k}\right)$= equation  \eqref{eq: gradient_Z}.
				\STATE \textbf{for} $k=1, \ldots, K$ \textbf{do}
				 \STATE \quad  $\boldsymbol{U}^{k}=\boldsymbol{Z}^{k}-\frac{1}{C}\nabla \mathcal{L}\left(\boldsymbol{Z}^{k}\right)$.
				 \STATE \textbf{end for}
				 \STATE $\Bar{\boldsymbol{U}}\leftarrow$ \textbf{Transform} ($\boldsymbol{U}^{k}$).
				 \STATE \textbf{Set} $\boldsymbol{A}_{0}=\Bar{\boldsymbol{U}}$,  $\boldsymbol{M}_n=0$,   $\boldsymbol{Q}_n=0$.
					\STATE \textbf{for} $n=0, \ldots, N$ \textbf{do}
					\STATE \quad $\boldsymbol{V}_{(n)}\leftarrow\operatorname{prox}_{\mathcal R 2}\left(\boldsymbol{A}_{(n)}+\boldsymbol{M}_{(n)}\right)$, see equation \eqref{eq:R_2}.
					\STATE \quad $\boldsymbol{M}_{(n+1)}\leftarrow\boldsymbol{A}_{(n)}+\boldsymbol{M}_{(n)}-\boldsymbol{V}_{(n)}$.
					\STATE \quad $\boldsymbol{A}_{(n+1)}\leftarrow\operatorname{prox}_{\mathcal R_1}\left(\boldsymbol{V}_{(n)}+\boldsymbol{Q}_{(n)}\right)$, see equation \eqref{eq:lasso}.
					\STATE \quad $\boldsymbol{Q}_{(n+1)}\leftarrow\boldsymbol{V}_{(n)}+\boldsymbol{Q}_{(n)}-\boldsymbol{A}_{(n+1)}$.
                    \STATE \quad\textbf{Break}: if $$\left\|\boldsymbol{A}_{(n+1)}-\boldsymbol{A}_{(n)}\right\|_F<\epsilon.$$
                \STATE \textbf{end for}
                \STATE $\Bar{\boldsymbol{U}}\leftarrow$ $\boldsymbol{A}_{({N}+1)}$.
                \STATE $\tilde{\boldsymbol{Z}} \leftarrow$ \textbf{Inverse-Transform} ($\Bar{\boldsymbol{U}}$).
                \STATE \textbf{return} $\tilde{\boldsymbol{Z}}$.
			\end{algorithmic}
		\end{algorithm}
\end{figure}

Due to the acyclicity term $h(\boldsymbol{Z}^{k})$, we are not able to derive a closed-form solution for equation  \eqref{eq:Z_k_hat}.
To tackle this issue, we utilize a proximal gradient method which constructs a quadratic approximation of $\mathcal{L}(\tilde{\boldsymbol{Z}})$ at $\tilde{\boldsymbol{Z}}_{(t-1)}$, and updates $\tilde{\boldsymbol{Z}}_{(t)}$ by 

\begin{equation}
\tilde{\boldsymbol{Z}}_{(t)}=\underset{\tilde{\boldsymbol{Z}}}{\arg \min }\{\mathcal{L}(\tilde{\boldsymbol{Z}}_{(t-1)})+\sum_{k=1}^{K}\left\langle\boldsymbol{Z}^{k}-\boldsymbol{Z}^{k}_{(t-1)}, \nabla \mathcal{L}_{k}(\boldsymbol{Z}^{k}_{(t-1)})\right\rangle +\frac{C}{2}\sum_{k=1}^{K}\left\|\boldsymbol{Z}^{k}-\boldsymbol{Z}^{k}_{(t-1)}\right\|_F^2+\mathcal{R}(\tilde{\boldsymbol{Z}})\},
\label{eq: tilde_Z}
\end{equation}
where $\nabla \mathcal{L}_k(\cdot)$ is the gradient of $\mathcal{L}_k(\cdot)$, $C>0$ is the Lipschitz constant of $\nabla \mathcal{L}_k(\cdot)$.
After some simple manipulations of equation \eqref{eq: tilde_Z}, e.g., ignoring constant terms of $\boldsymbol{Z}_{(t-1)}^{k}$, we have
\begin{equation}
\begin{aligned}
\tilde{\boldsymbol{Z}}_{(t)}&=\underset{\tilde{\boldsymbol{Z}}}{\arg \min }\{\frac{C}{2}\sum_{k=1}^{K}\left\|\boldsymbol{Z}^{k}-\left(\boldsymbol{Z}_{(t-1)}^{k}-\frac{1}{L}\nabla \mathcal{L}_{k}\left(\boldsymbol{Z}_{t-1}^{k}\right)\right)\right\|_F^2+\mathcal{R}(\tilde{\boldsymbol{Z}})\},\\
&=\operatorname{prox}_{C \mathcal{R}}\left(\sum_{k=1}^{K}\left(\boldsymbol{Z}_{(t-1)}^{k}-\frac{1}{L}\nabla \mathcal{L}_{k}\left(\boldsymbol{Z}_{(t-1)}^{k}\right)\right)\right),
\label{eq: proximal_operator}
\end{aligned}
\end{equation}
where
\begin{equation}
\nabla \mathcal{L}_{k}\left(\boldsymbol{Z}_{(t-1)}^{k}\right)=\alpha_{k,(t-1)} \nabla h(\boldsymbol{Z}^{k}_{(t-1)})+\rho_{1,(t-1)}h(\boldsymbol{Z}^{k}_{(t-1)})\nabla h(\boldsymbol{Z}^{k}_{(t-1)})-\beta_{k,(t-1)}
+\rho_{2,(t-1)}(\boldsymbol{Z}^{k}_{(t-1)}-\boldsymbol{W}^{k}),
\label{eq: gradient_Z}
\end{equation}    
and $\operatorname{prox}_g(v)=\underset{x }{\arg \min }(\frac{1}{2}\|x-v\|_{F}^2+g(x))$ denotes the proximal operator.

Unlike the calculation of $\boldsymbol{W}^{k}$, we cannot separate the equation \eqref{eq: proximal_operator} by $k$ because of the grouped constraint $\mathcal{R}(\tilde{\boldsymbol{Z}})$.
Instead, we must estimate the whole set of matrices $\tilde{\boldsymbol{Z}}_{(t)}$ jointly.
Inspired by \cite {gibberd2017regularized}, we use the following iterative proximal projection step to solve.

Let  $\boldsymbol{U}^{k}=\boldsymbol{Z}_{(t-1)}^{k}-\frac{1}{C}\nabla \mathcal{L}_{k}\left(\boldsymbol{Z}_{(t-1)}^{k}\right)$, and perform  transform procedure (see Appendix\ref{sec:global_update}) for $\boldsymbol{U}^{k} \rightarrow \Bar{\boldsymbol{U}}$.
Then, equation \eqref{eq: proximal_operator} can be re-written as
\begin{equation}
\underset{\Bar{\boldsymbol{Z}}}{\min }\underbrace{\frac{1}{2}\left\|\Bar{\boldsymbol{Z}}-\Bar{\boldsymbol{U}}\right\|_F^2}_{\mathcal{L}(\Bar{\boldsymbol{Z}})}+\underbrace{\lambda_{1}\|\Bar{\boldsymbol{Z}}\|_{1}}_{\mathcal{R}_1(\Bar{\boldsymbol{Z}})}+\underbrace{\lambda_{2}  \|\boldsymbol{D}\Bar{\boldsymbol{Z}}\|_{2,1}}_{\mathcal{R}_2(\Bar{\boldsymbol{Z}})},
\label{eq: fused}
\end{equation}
where $\boldsymbol{D} \in \mathbb{R}^{(K-1) \times K}$ is a backwards difference matrix of the form $D_{i, i}=-1, D_{i, i+1}=1 \text { for } i=1, \ldots, K-1$, and $\|A\|_{2,1}\text { := }\sum_k\left\|A_{k, \cdot}\right\|_2$ denotes the group $\ell_{2,1}$ norm. 
Actually, equation \eqref{eq: fused} is equivalent to the Group-Fused Lasso Signal Approximator (GFLSA) defined by  \cite{gibberd2017regularized}.
We denote the objective in equation \eqref{eq: fused} by $f\left(\Bar{\boldsymbol{U}}; \lambda_1, \lambda_2\right)=\operatorname{prox}_{ \mathcal{R}_{1}+\mathcal{R}_{2}}\left(\Bar{\boldsymbol{U}}\right)$.
Then, we adopt the Dykstra-like iterative proximal algorithm (DIPA)  \citep{combettes2011proximal} to handle $\operatorname{prox}_{ \mathcal{R}_{1}+\mathcal{R}_{2}}\left(\Bar{\boldsymbol{U}}\right)$ and find a feasible solution for both the group fused lasso penalty and the lasso penalty.
We summarize the details of iteration in Algorithm \ref{algorithm: DIPA}.

\begin{figure}[]
		\renewcommand{\algorithmicrequire}{\textbf{Input:}}
		\renewcommand{\algorithmicensure}{\textbf{Output:}}
	    %\removelatexerror
		\begin{algorithm}[H]
			\caption{ ADMM of NOTEARS-PFL}
			\begin{algorithmic}\label{algorithm: ADMM}
				\REQUIRE $K$: Number of site; $\mathbf{x}^{k}$: Original data; $\tilde{\boldsymbol{Z}}_{(0)}$: Initial  global  adjacency matrices; $\rho_{1,0}$, $\rho_{2,0}$, $\alpha_{1,0}, \ldots, \alpha_{K,0}$, $\beta_{1,0}, \ldots, \beta_{K,0}$: Initial parameters; $T$: The maximum of iterations; $\epsilon$: Tolerance.
					\ENSURE
		      $\hat{\boldsymbol{Z}^{k}}$: Estimated adjacency matrix.
				\STATE \textbf{for} iterations $t=1$ \textbf{to} $T$ \textbf{do}
				 \STATE \quad\textbf{Local update step}:
					\STATE \qquad\textbf{for} site $k=1, \ldots, K$ \textbf{do}
					\STATE \quad\qquad {Update $\boldsymbol{W}^{k}_{(t)}$ according to equation \eqref{eq:W_k_hat}.}
					\STATE \qquad\textbf{end for}
					 \STATE \quad\textbf{Global update step}:
					\STATE \qquad\textbf{for} variables $\tilde{\boldsymbol{Z}}_{(t-1)}$, $\boldsymbol{\Theta}_{(t-1)}$ in the center \textbf{do}
					\STATE \quad\qquad $\boldsymbol{Z}^{k}_{(t)}\leftarrow$ \textbf{DIPA} ($\boldsymbol{W}^{k}_{(t)}, \boldsymbol{Z}^{k}_{(t-1)}$).
					\STATE \quad\qquad Update $\boldsymbol{{\tilde{\Theta}}}_{(t-1)}$ according to  equation \eqref{eq: Theta}.
					\STATE \qquad\textbf{end for}
                    \STATE \quad\textbf{Break}: if $$\sum_{k=1}^K \frac{\left\|\boldsymbol{Z}^{k}_{(t)}-\boldsymbol{Z}^{k}_{(t-1)}\right\|_F}{\left\|\boldsymbol{Z}^{k}_{(t)}\right\|_2}<\epsilon.$$
                \STATE \textbf{end for}
                \STATE \textbf{return} $\tilde{\boldsymbol{Z}}=(\boldsymbol{Z}^{1}_{(T)},...,\boldsymbol{Z}^{K}_{(T)})$.
			\end{algorithmic}
		\end{algorithm}
\end{figure}

\subsubsection{Federated Global Update of $\tilde{\boldsymbol{\Theta}}$}
The update of $\tilde{\boldsymbol{\Theta}}$ follows the following iterative steps
\begin{equation}
\begin{aligned}
\beta_{k,t}&:=\beta_{k,t-1}+\rho_{2,t-1}\left(\boldsymbol{W}^{k}_{(t)}-\boldsymbol{Z}^{k}_{(t)}\right),\\
\alpha_{k,t}&:=\alpha_{k,t-1}+\rho_{1,t-1} h\left(\boldsymbol{Z}^{k}_{(t)}\right),\\
\rho_{1,t}&:=\gamma_{1}\rho_{1,t-1},\\
\rho_{2,t}&:=\gamma_{2}\rho_{2,t-1},\\
\end{aligned}
\label{eq: Theta}
\end{equation}
where $\gamma_{1},\gamma_{1} \in \mathbb{R}$  are hyperparameters that respectively control the increasing speeds of the coefficients $\rho_{1}, \rho_{2}$. 
The overall ADMM algorithm of NOTEARS-PFL is thereby given in Algorithm \ref{algorithm: ADMM}.

\section{Simulations}\label{sec:experiments}
\subsection{Simulation Design of Synthetic Data}
We simulate synthetic data according to equation \eqref{eq:Xi}.
First, we generate a random graph $G_{truth}$ based on  Erdös–Rényi (ER) model \citep{erdHos1960evolution}. 
To obtain a set of related graphs, we apply perturbations to $G_{truth}$ to create 
$K$ similar but different graphs $\tilde{G}= \{G^{1}, \ldots, G^{K}\}$.
The details of the definition of perturbation are shown in Appendix \ref{sec:synthetic_data}, and we denote the perturbation level as $p_l$.

Given $\tilde{G}$, we assign uniformly random edge weights to obtain $K$ weights matrices $\tilde{\boldsymbol{W}}=\{\boldsymbol{W}^{1}, \ldots, \boldsymbol{W}^{K}\}$, where the non-zero entries are sampled uniformly at random form $[-2,-0.5] \cup[0.5,2]$.
Given $\tilde{\boldsymbol{W}}$, we generate the datasets $\mathcal{D}=\{\mathbf{x}^{1},\dots,\mathbf{x}^{K}\}$ based on equation \eqref{eq:Xi}
with standard Gaussian noise. 

To evaluate and compare the performance of the proposed method, we apply the NOTEARS \citep{zheng2018dags} method to estimate each network independently from each site’s local data (NOTEARS-SIG for short), and apply the NOTEARS method to learn a common network structure for all contexts (NOTEARS-AVG for short), \emph{i.e.}, an “average” network that treats all site’s data as samples in one dataset.
We also compare the performance between NOTEARS-ADMM \citep{ng2022towards} and the proposed method.

We use the following four metrics to evaluate the learning performance of different methods (see Appendix \ref{sec:synthetic_data} for a detailed definition of metrics). 
\begin{itemize}
    \item \textbf{Edge arrowhead error:}
     \begin{equation}
     \text {\emph{Error}}=\frac{FP+F N}{T P+T N+F P+F N}. \label{eq:error}
    \end{equation}
    \item \textbf{Edge arrowhead  precision:}
    \[
     \text { \emph{Precision} }=\frac{T P}{T P+F P}.
    \]
     \item \textbf{Edge arrowhead  F-score:}
     \[
      \text { \emph{F-score}}=2 \cdot \frac{\text { \emph{Precision} } \cdot \text {\emph{Recall} }}{\text { \emph{Precision} }+\text {\emph{Recall}}},
     \]
     where $\text { \emph{Recall}}=TP/(TP+FN)$.
     \item \textbf{Edge arrowhead structural Hamming distance (SHD):} Given two structures, this measure is the minimum number of edges needed to convert the learned graph into the true one.
\end{itemize}

We compare the performance of the proposed NOTEARS-PFL, NOTEARS-SIG, NOTEARS-AVG, and NOTEARS-ADMM on the following three aspects. 

\textbf{Performance vs. the number of variable:} We fix the number of sites $K=10$, perturbation level $p_l=10\%$, set the total sample size $n_{t}=\sum_{k=1}^{K}n_{k}$ equals to three times the number of variable $d$ ($n_{t}=3d$, $n_{k}=0.3d$), and compare the model performance of network structure learning for different variable number $d=10, 20, 30, 40, 50, 60, 70, 80$.

\textbf{Performance vs. the number of site:} We fix the number of variable $d=50$, perturbation level $p_l=10\%$.
We set the total sample size $n_{t}=256$, and distributed evenly across the number of site $K=2, 4, 8, 16, 32, 64$.

\textbf{Performance vs. perturbation level:} We fix the number of site $K=10$, the number of variable $d=30$, sample size $n_t=3d$, and vary the perturbation level $p_l=5\%, 10\%, 15\%, 20\%, 30\%$.

The evaluation metrics for one simulation run are averaged over the $K$ sites' dataset, and we repeat each experiment $10$ times. The final evaluation metrics are thereby averaged over the ten simulation runs.

\begin{figure}[h]
     \centering
  {\includegraphics[width=0.7\linewidth]{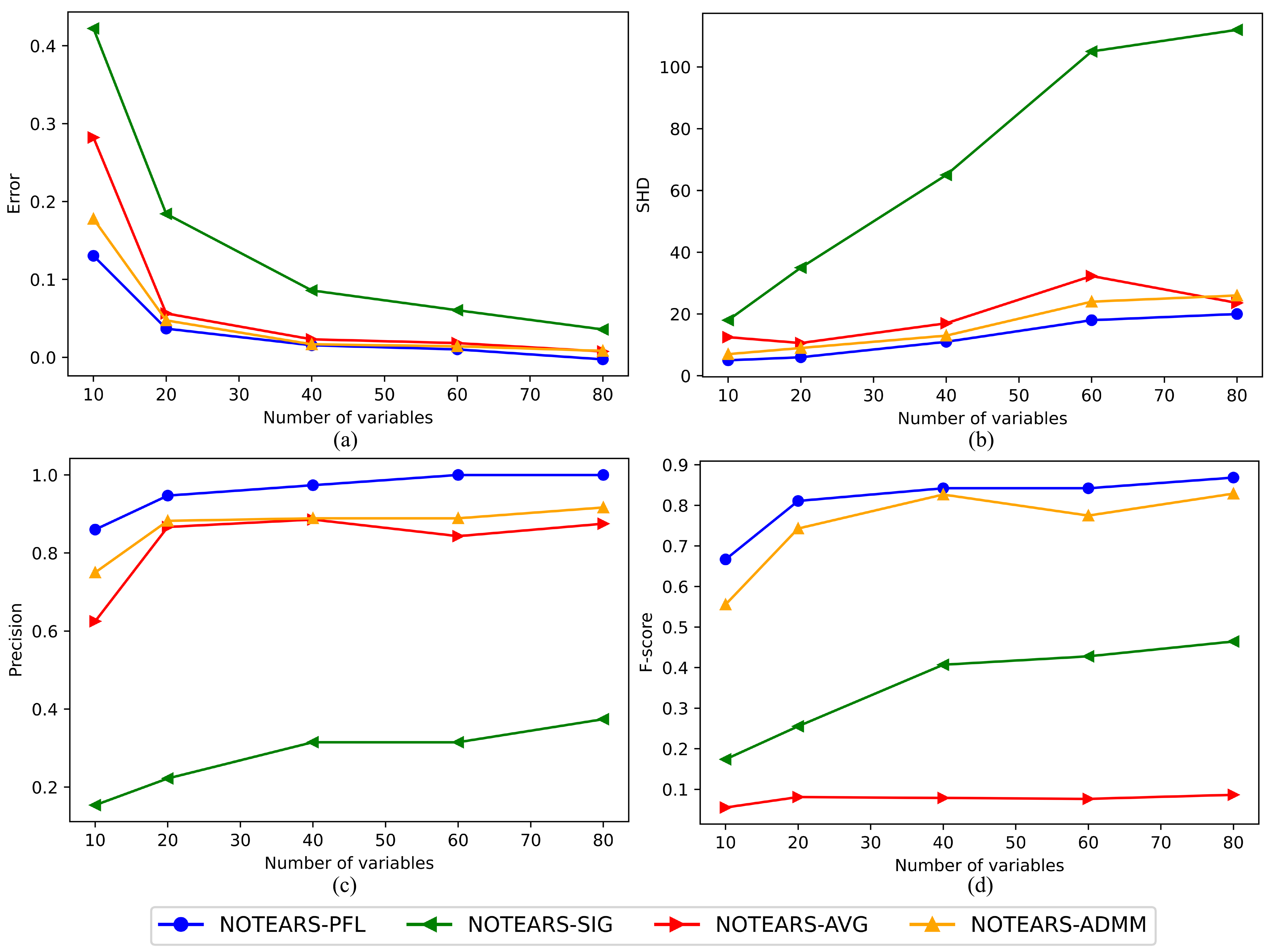}}
  \caption{Simulation results for synthetic data under different variable numbers. (a) Average edge arrowhead error (lower is better); (b) Average edge arrowhead SHD (lower is better); (c) Average edge arrowhead precision (higher is better); (d) Average edge arrowhead F-score (higher is better).}
  \label{fig: preformance_variable_number}
\end{figure}

\begin{figure}
     \centering
  {\includegraphics[width=0.7\linewidth]{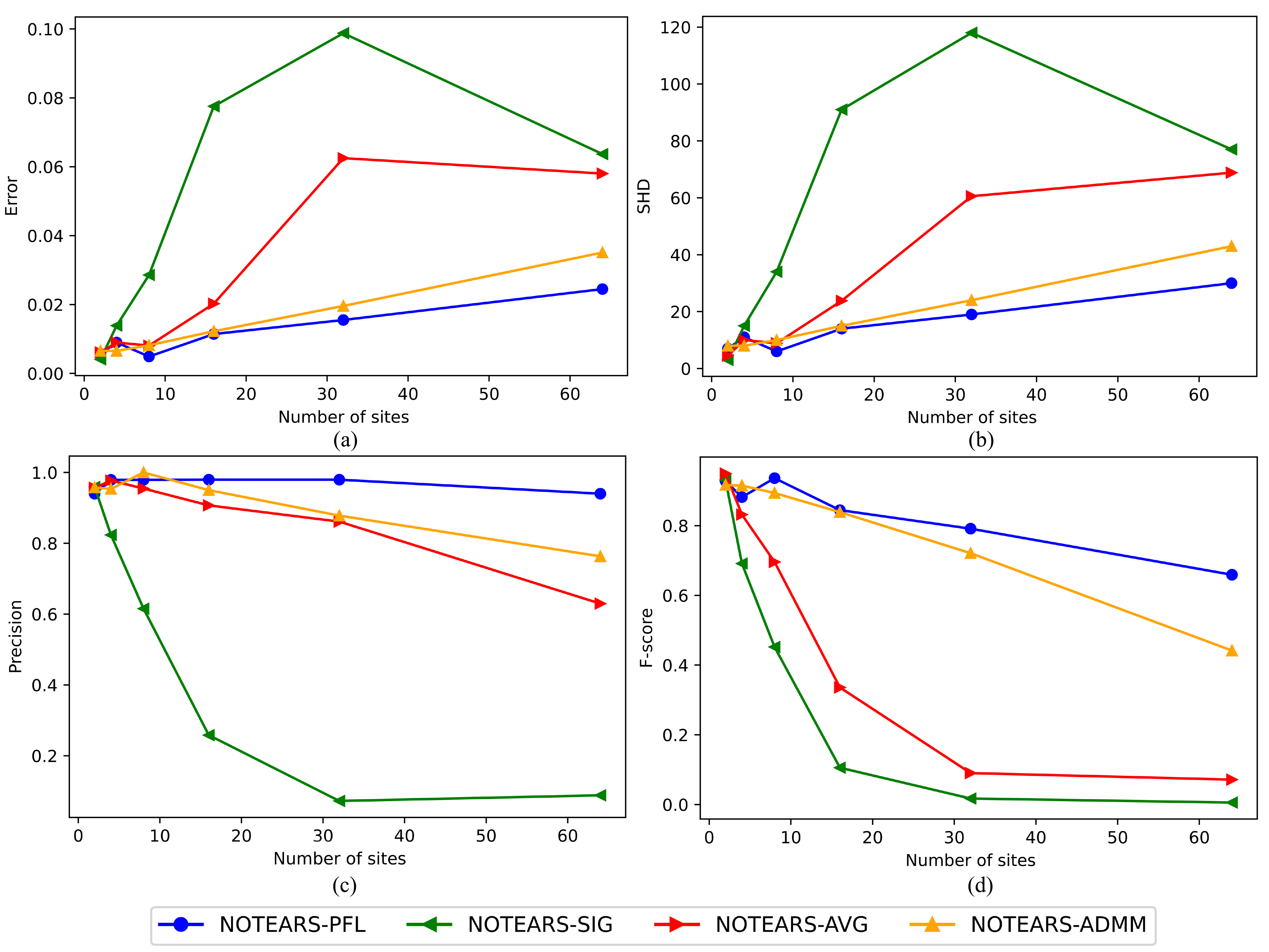}}
  \caption{Simulation results for synthetic data under different site numbers. (a) Average edge arrowhead error (lower is better); (b) Average edge arrowhead SHD (lower is better); (c) Average edge arrowhead precision (higher is better); (d) Average edge arrowhead F-score (higher is better).}
  \label{fig: preformance_client_number}
\end{figure}

\begin{figure}
     \centering
  {\includegraphics[width=0.7\linewidth]{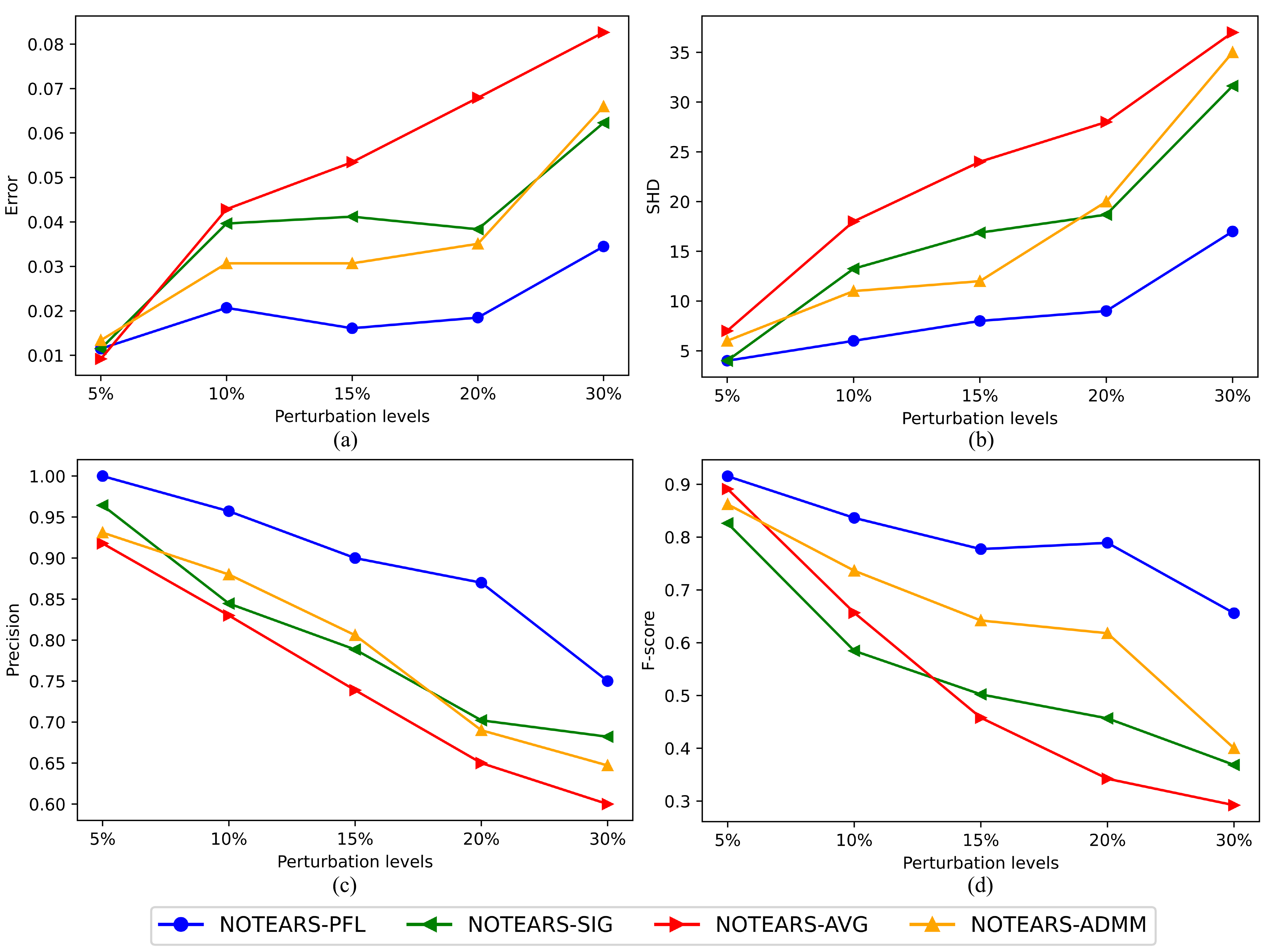}}
  \caption{Simulation results for synthetic data under different perturbation levels (a) Average edge arrowhead error (lower is better); (b) Average edge arrowhead SHD (lower is better); (c) Average edge arrowhead precision (higher is better); (d) Average edge arrowhead F-score (higher is better).}
  \label{fig: preformance_perturbation_level}
\end{figure}

In the synthetic data experiment, we first evaluate the effect of the number of variables $d$ on the performance of each compared method. The results are shown in  Figure \ref{fig: preformance_variable_number}.
It is clear that NOTEARS-PFL performs the best among all the methods.
The average arrowhead adjacency error of NOTEARS-PFL has a relatively low level ranging from $0.02$ to $0.13$. In contrast, for NOTEARS-SIG, the average arrowhead adjacency error is far larger than NOTEARS-PFL, ranging from $0.04$ to $0.43$.
This may be because the sample size of a single dataset is too small for NOTEARS-SIG to give a satisfactory estimate. 
The same result is observed in the average arrowhead SHD.
For the other two metrics, NOTEARS-PFL consistently results in higher average arrowhead precision and F-score than the other three methods.
Compared with NOTEARS-SIG and NOTEARS-AVG, the average arrowhead precision of NOTEARS-PFL is also more stable with respect to different variable numbers, indicating that it can successfully identify most of the edges in high-dimensional settings.

We also test the effect of the number of sites $K$ on the performance of each method. The results are shown in 
Figure \ref{fig: preformance_client_number}. Clearly,  NOTEARS-PFL performs better than NOTEARS-SIG, NOTEARS-AVG, and NOTEARS-ADMM for all the tested cases. 
As $K$ increases, the performance of NOTEARS-SIG, NOTEARS-AVG decline rapidly.
Although the performance of NOTEARS-PFL also decreases with the increase of $K$, it still provides a relatively stable and satisfying performance.
As expected, NOTEARS-SIG does not perform better at larger $K$.
A possible reason is that NOTEARS-SIG only utilizes the information of a single dataset whose sample size decreases with the increase of $K$.
On the contrary, NOTEARS-PFL works well in the setting with a large site number because information can be exchanged during optimization.
These results are also consistent with the existing literature \citep{ng2022towards}.

Finally, we examine the effect of the perturbation level $p_l$ on the performance of each method. The results are shown in Figure \ref{fig: preformance_perturbation_level}.
It is clear that NOTEARS-PFL significantly outperforms the other three methods.
It can be seen that with the increase of perturbation level, average arrowhead error and SHD increase significantly, while average arrowhead precision and F-score decrease.
When the perturbation level is relatively low, the difference between the four methods is relatively small, but when the perturbation level is relatively large, the advantage of NOTEARS-PFL is more significant.

\section{Application to Real-world rs-fMRI Data}\label{sec: real_world}

\subsection{Data and Preprocessing}

\quad\;\;\textbf{(1) Data acquisition:} The multi-site rs-fMRI datasets in this study were obtained from the DecNef Project Brain Data Repository\footnote{https://bicr-resource.atr.jp/srpbsopen/}, collected as part of the Japanese Strategic Research Program for the Promotion of Brain Science (SRPBS)  database project \citep{tanaka2021multi}. 
This dataset consists of 255 MDD patients (135 men versus 120 women) and 791 HC participants (426 men versus 365 women) from 8 sites.
Some samples were discarded as some sites only have rs-fMRI data on HC participants.
As a result, 238  MDD patients and 475 HC participants from 5 sites were used in this study (see Table \ref{tab: summary}).
The demographic characteristics of all subjects in both datasets are summarized in Table \ref{tab:demographic}.

\textbf{(2) Data preprocessing :} Each subject of the data underwent a single rs-fMRI session and a structural MRI session. 
During the fMRI scan, participants were asked to relax, stay awake and think about anything in particular.
Detailed imaging parameters of rs-fMRI and T1-weighted (T1w) structural MRI at each site are shown in Table \ref{tab: imaging_protocol}.
We applied the following preprocessing steps to the data by using DPARSF \citep{yan2010dparsf}  software\footnote{http://www.rfmri.org/DPARSF}. 
For each participant, we removed the first 10 MRI volumes.
Slice-time correction and head motion were corrected on the remaining images. 
The T1-weighted images were unified segment segmentation and diffeomorphic anatomical registration.
The rs-fMRI data were smoothed with a 6mm full-width at half-maximum Gaussian kernel and the Friston 24-parameter model was used to regress head motion effects further.
Then, all the time series were filtered by linear detrending and a temporal band-pass filter (0.01 - 0.08 Hz).

\textbf{(3) Feature extraction:} After preprocessing, we estimated the time series of 116 ROIs based on the automated anatomical labeling (AAL) template \citep{tzourio2002automated}.
Details of cortical and subcortical regions of interest (ROIs) defined in the AAL template are presented in Table \ref{tab: roi} and Table \ref{tab: roi_continue}.
We selected 56 ROIs from 116 ROIs because these regions are considered to be related to MDD in the literature.
The names of the selected ROIs are shown in bold in Table \ref{tab: roi} and Table \ref{tab: roi_continue}.
The voxel-wise fMRI time courses were averaged into one regional average time course within each ROI. 
The data for each participant was then in the form of an $X\times$ 56 matrix ($X$ sampling points in each regional average fMRI time course and 56 selected ROIs).

\subsection{Experimental Setup} 
We randomly select 70\% of MDD patients and HC participants from each site and aggregate their rs-fMRI data into the MDD discovery dataset and HC discovery dataset for that site, respectively.
Similarly, we aggregate the remaining 30\% into the MDD validation dataset and HC validation dataset for that site, respectively.
We analyze the 5 sites' discovery datasets of MDD patients, which are regarded as 5 related tasks. 
Then we can obtain a brain connection network of MDD for each site. 
To study the differences in brain FC, the brain connection networks of HC for five sites, which are viewed as another set of related tasks, are also obtained using the proposed   NOTEARS-PFL method.
According to the estimated BN structures, we analyze the abnormal FC in MDD by the following aspects.

\subsubsection{Important Nodes Identification}
By identifying important nodes, we reveal the aberrant nodal characteristics of the brain functional connectivity networks of MDD from multi-site datasets. 

\subsubsection{Overlapping Connections Identification}
By identifying overlapping connections, we obtain the common functional connections for MDD patients from multiple sites.
By comparing with the overlapping connections of HC participants, we further investigate the altered functional connections in MDD patients, which provide the potential biomarkers for clinic treatment of MDD.

\subsubsection{Site-specific Connections Identification}
By identifying the site-specific connections in different sites, we investigate the specific FC alterations in MDD at each site due to the differences in image acquisition protocols and patient populations at different sites.

\subsubsection{Comparison with Other Methods}
We randomly select 5 MDD patients and 5 HC participants from each site's MDD validation dataset and HC validation dataset and repeat such selection 10 times.
We then have 10 datasets containing MDD data and HC data from the 5 sites.
Based on these datasets, we compare NOTEARS-PFL with  NOTEARS-SIG and existing standard analysis methods of rs-fMRI data: GES \citep{chickering2002optimal}, PC \citep{spirtes1991algorithm}, and GC \citep{granger1969investigating}. 
To further test whether NOTEARS-PFL is able to learn the differences between MDD patients and HC participants.

\subsection{Experimental Results}
In this section, we use NOTEARS-PFL method to analyze the multi-site rs-fMRI data from 5 sites.
We can obtain the brain functional connection network of MDD patients for each site.
To study the differences in brain FC, we also obtain the brain functional connection network of HC participants for each site.

\subsubsection{Results of Important Nodes}
To better understand the differences in brain FC between MDD patients and HC participants through the estimated directed brain networks from multi-site rs-fMRI data, we further use the connection degree to identify some important nodes.
The connection degree (including in and out degree) is a commonly used measure of functional connectivity \citep{liu2022learning}.
More specifically, it represents the amount of directed functional connections. The greater the connection degree, the higher the effective FC. 
Identifying important nodes further helps to discover the nodes associated with the disease.
Table \ref{tab: connection_degree} shows the differences in total connection degrees between MDD patients and HC participants at 5 sites.
The total out connection degrees in Table \ref{tab: connection_degree} are 657 for MDD and 478 for HC, indicating that MDD has 37.4\% effective FC than HC.
Similarly, MDD also has 23.4\% more effective FC than HC, with a total in connection degrees of 516 and 418, respectively.
The rise of FC is an apparent robust biomarker of MDD disease \citep{zamoscik2014increased}, which can also be found in our results.

\begin{table}[h]
\centering
\caption{The top 10 out and in connection degrees of MDD patients and HC participants}
\label{tab: connection_degree}
\setlength{\tabcolsep}{3pt}
\resizebox{0.9\linewidth}{!}{
\begin{tabular}{@{}cccccccccc@{}}
\toprule
                   & Index & ROI                                    & Abbreviation & Degree &                   & Index & ROI                                    & Abbreviation & Degree \\ \midrule
\textbf{MDD (Out)} & 77    & Thalamus                               & THA.L        & 75     & \textbf{HC (Out)} & 55    & Fusiform gyrus                         & FFG.L        & 58     \\
                   & 29    & Insula                                 & INS.L        & 74     &                   & 37    & Hippocampus                            & HIP.L        & 57     \\
                   & 35    & Cingulate gyurs, posterior part        & PCG.L        & 72     &                   & 31    & Cingulate gyrus, anterior part         & ACG.L        & 55     \\
                   & 33    & Cingulate gyrus, mid part              & DCG.L        & 68     &                   & 38    & Hippocampus                  & HIP.R       & 49     \\
                   & 34    & Cingulate gyrus, mid part              & DCG.R        & 67     &                   & 9     & Middle frontal gyrus, orbital          & HIP.R        & 47     \\
                   & 31    & Cingulate gyrus, anterior part         & ACG.L        & 63     &                   & 29    & Insula                                 & INS.L        & 46     \\
                   & 72    & Caudate                                & CAU.R        & 62     &                   & 77    & Thalamus                               & THA.L        & 45     \\
                   & 71    & Caudate                                & CAU.L        & 61     &                   & 33    & Cingulate gyrus, mid part              & DCG.L        & 43     \\
                   & 55    & Fusiform gyrus                         & FFG.L        & 58     &                   & 56    & Fusiform gyrus                         & FFG.R        & 40     \\
                   & 79    & Heschl gyrus                           & HES.L        & 57     &                   & 79    & Heschl gyrus                           & HES.L        & 38     \\
Total              &       &                                        &              & 657    & Total             &       &                                        &              & 478    \\
\textbf{MDD (In)}  & 24    & Superior frontal gyrus, medial         & SFGmed.R     & 79     & \textbf{HC (In)}  & 3     & Superior frontal gyrus, dorsolateral   & SFGdor.L     & 55     \\
                   & 90    & Inferior temporal gyrus                & ITG.R        & 62     &                   & 9     & Middle frontal gyrus, orbital          & ORBmid.L     & 48     \\
                   & 26    & Superior frontal gyrus, medial orbital & ORBsupmed.R  & 59     &                   & 26    & Superior frontal gyrus, medial orbital & ORBsupmed.R  & 45     \\
                   & 89    & Inferior temporal gyrus                & ITG.L        & 54     &                   & 90    & Inferior temporal gyrus                & ITG.R        & 44     \\
                   & 3     & Superior frontal gyrus, dorsolateral   & SFGdor.L     & 48     &                   & 10    & Middle frontal gyrus, orbital          & ORBmid.R     & 41     \\
                   & 46    & Cuneus                                 & CUN.R        & 46     &                   & 14    & Inferior frontal gyrus, triangular     & IFGtriang.R  & 40     \\
                   & 11    & Inferior frontal gyrus, opercular      & IFGoperc.L   & 45     &                   & 24    & Superior frontal gyrus, medial         & SFGmed.R     & 39     \\
                   & 25    & Superior frontal gyrus, medial orbital & ORBsupmed.L  & 42     &                   & 1     & Precentral gyrus                       & PreCG.L      & 38     \\
                   & 41    & Amygdala                               & AMYG.L       & 41     &                   & 89    & Inferior temporal gyrus                & ITG.L        & 36     \\
                   & 32    & Cingulate gyrus, anterior part         & ACG.R        & 40     &                   & 6     & Superior frontal gyrus, orbital        & ORBsup.R     & 32     \\
Total              &       &                                        &              & 516    & Total             &       &                                        &              & 418    \\ \bottomrule
\end{tabular}
}
\end{table}

From Table \ref{tab: connection_degree}, we can find some ROI nodes with high connection degrees, such as THA.L (thalamus), INS.L (insula), SFGmed.R (superior frontal gyrus, medial) in MDD, and FFG.L (fusiform gyrus) and HIP.L (hippocampus) in HC.
Specifically, THA.L and INS.L have more outgoing connections in MDD than in HC.
It indicates that THA.L and INS.L of MDD have a higher essential impact on other ROI nodes than HC.
The result from NOTEARS-PFL is in line with literature \citep{kang2018functional, avery2014major,porta2021multimetric}.
According to \cite {kang2018functional}, emerging evidence indicates that the enhanced thalamus FC is an important feature of the underlying pathophysiology of MDD.
This abnormal FC is related to the core clinical symptoms of MDD, such as anxiety, memory, and attention.
Our result further implies that the impaired FC of the hippocampus is potentially an important biomarker of MDD.
Compared to HC, the bilateral hippocampus in MDD lack FC.
As the core region of the limbic system, the hippocampus plays an important role in regulating motivation and emotion.
The mode of emotion regulation in MDD may be related to the decreased FC of the hippocampus \citep{cao2012disrupted}.
SFGmed.R (superior frontal gyrus, medial) and ITG.R (inferior temporal gyrus) have more incoming connections than outgoing connections, which indicates that SFGmed.R and ITG.R are mainly affected by other ROI nodes.
These findings are also consistent with the physiological discoveries of MDD disease \citep{porta2021multimetric}.

\subsubsection{ Results of Overlapping Connections}

\begin{figure}[h]
     \centering
  {\includegraphics[width=0.8\textwidth]{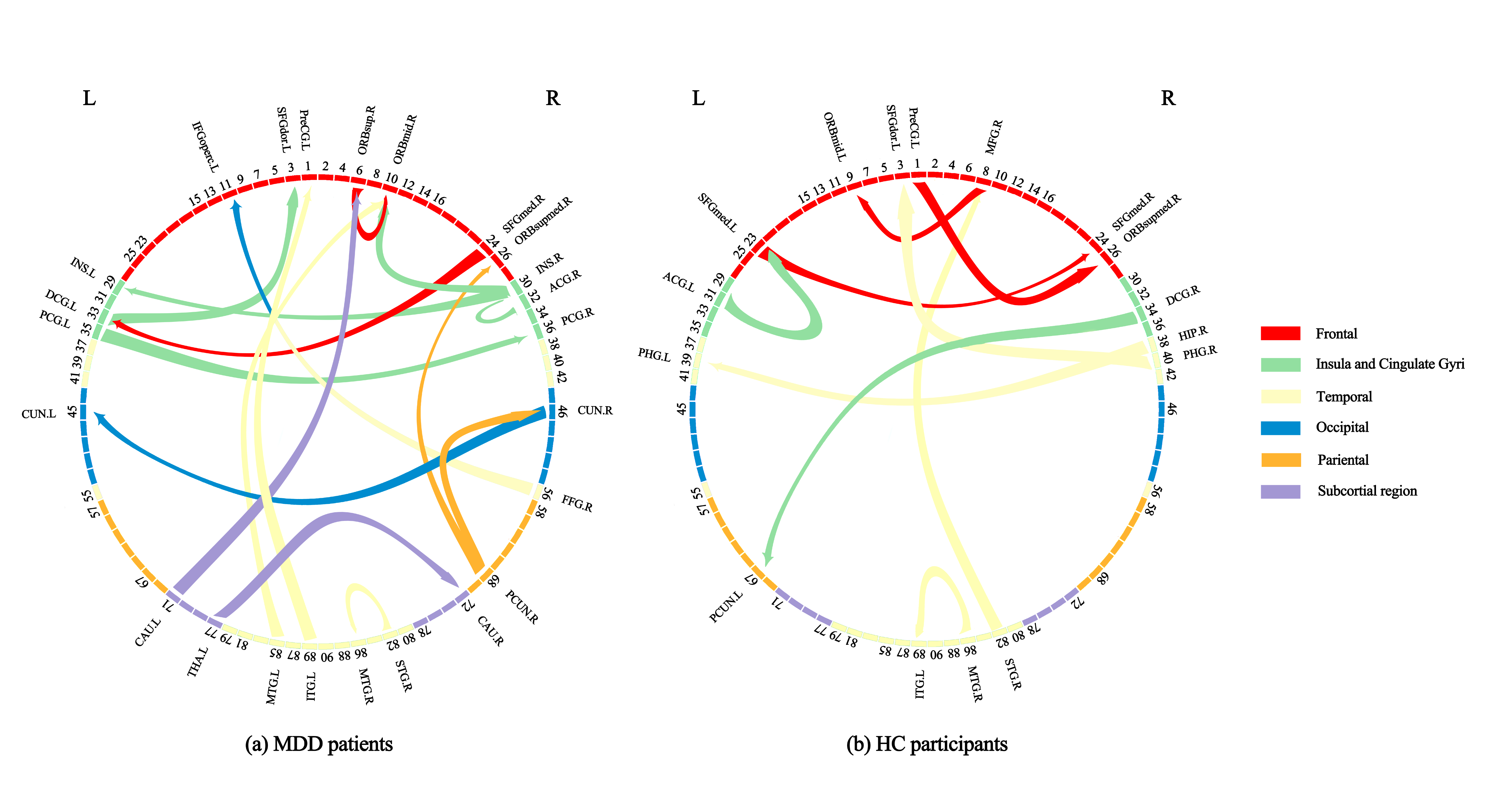}}
  \caption{The overlapping connections between the directed networks of 5 sites of MDD patients and HC participants. } 
  \label{fig: connection}
\end{figure}

Figure \ref{fig: connection} depicts the overlapping connections between the directed networks of 5 sites of MDD patients and HC participants.
The overlapping connection refers to the connection shared by five sites.
We can use it to study the universal connection between MDD and compare the differences between MDD and HC.
From Figure \ref{fig: connection}, we can find some abnormal functional connections in MDD: INS.R (insula) $\rightarrow$ ORBmid.R (middle frontal gyrus, orbital), INS.R (insula) $\rightarrow$ INS.L (insula), INS.R (insula)  $\rightarrow$ ACG.R (cingulate gyrus, anterior part), PCG.L (cingulate gyurs, posterior part) $\rightarrow$ PCG.R (cingulate gyurs, posterior part), THA.L (thalamus) $\rightarrow$ CAU.R (caudate), 
STG.R (superior temporal gyrus) $\rightarrow$ MTG.R (middle temporal gyrus).
Three common functional connections are associated with the insula, which further demonstrates the important role of the insula in identifying abnormal functional connections in MDD.
The insula has been shown to be a potential primary biomarker for the diagnosis of MDD \citep{mcgrath2013toward}.
In MDD patients,  the insula show increased FC with other limbic or paralimbic structures, particularly with the ventromedial prefrontal cortex (vmPFC) and  orbitofrontal cortex (OFC) \citep{avery2014major,drevets2008brain}, which is also illustrated in our results. 

\begin{figure}[h]
     \centering
  {\includegraphics[width=0.8\textwidth]{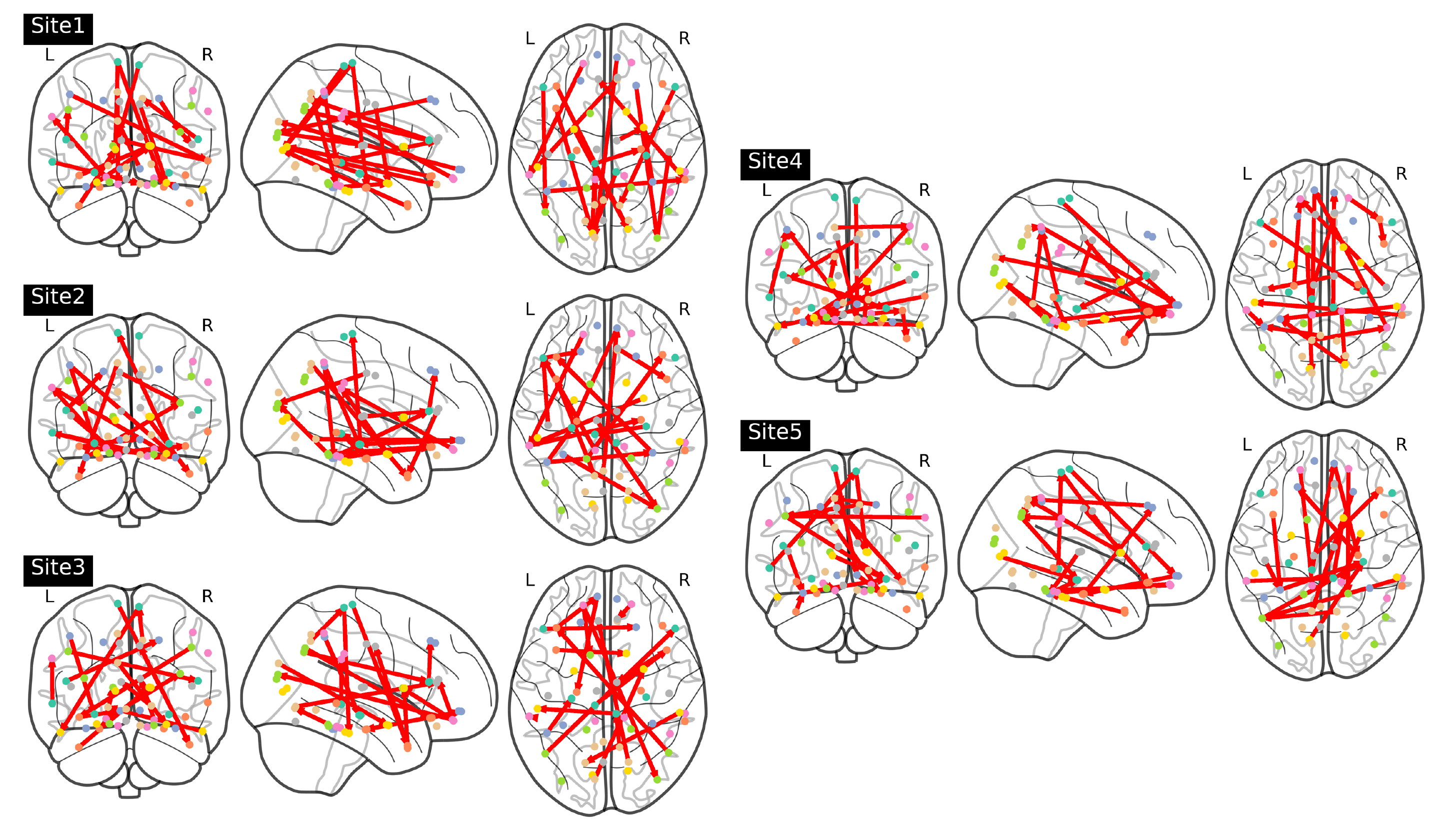}}
  \caption{ The site-specific connections between the directed networks of 5 sites of MDD patients. Site 1: Center of Innovation in Hiroshima (COI), Site 2: Kyoto University (KUT), Site 3: University of Tokyo (UTO), Site 4: Hiroshima Kajikawa Hospital (HKH), Site 5: Hiroshima University Hospital (HUH).}
  \label{fig: site}
\end{figure}

In addition to the abnormal insula structure in MDD patients, we also observe the  abnormalities of FC in posterior cingulate gyurs, thalamus, inferior temporal gyrus, and middle temporal gyrus, which are also consistent with findings in the works of literature \citep{khundakar2009morphometric, zhao2014brain}.
These cortices belong to the default mode network (DMN) and are considered to contribute greatly to MDD \citep{gong2015depression}.
The abnormalities of FC in DMN have been reported in many MDD studies \citep{vasudev2018bold}, especially the frontal lobe and temporal lobe. 
These lobes are suggested to be closely associated to the cognitive and information-processing abilities of MDD patients \citep{brzezicka2013integrative}.

\subsubsection{ Results of Site-specific Connections}
In this study, we are also interested in the connections specific to a site.
A site-specific connection is a connection that only exists on one site.
By identifying the site-specific connections in different sites, we can better understand the specificity of FC in MDD patients due to the  differences in image acquisition protocols and patient populations at different sites.
The site-specific connections of 5 sites of MDD patients are visualized in Figure \ref{fig: site}.
In the results of site 1, we can find the increased functional connection from frontal to pariental  and from frontal to temporal, especially the increased afferent connections to  precuneus.
The precuneus is considered one of the hubs of  DMN, which is generally associated with the rumination of negative and sad thoughts in MDD \citep{cheng2018functional}. It has been shown to play a key role in identifying MDD \citep{peng2015dissociated}.
A study of effective FC in MDD has shown a significant increase in forward connectivity from the middle and inferior temporal cortical areas to the precuneus \citep{ongur2003architectonic}, which is consistent with our study (the  connections IFGoperc.L $\rightarrow$ PCUN.R, IFGtriang.L $\rightarrow$ PCUN.L).

Similarly, in site 2, we find the enhanced FC within the frontal cortex, particularly the connections associated with the orbitofrontal cortex (the connections ORBsup.L $\rightarrow$ ORBsupmed.L, MFG.L $\rightarrow$ ORBmid.R, AMYG.L  $\rightarrow$ ORBmid.L).
The increased FC in the  orbitofrontal cortex in MDD patients is associated with emotionally negative self-perception \citep{cheng2016medial}.
The orbitofrontal cortex is also a part of the brain’s affective network (AN), and the increased functional connections of the orbitofrontal cortex in MDD have been consistently reported \citep{townsend2010fmri}.
The different effective FC patterns of the orbitofrontal cortex can also be used to reveal different pathophysiological mechanisms of different depression types or groups.

\subsubsection{ Comparison Results with Other Methods}
To further test whether NOTEARS-PFL is able to learn the differences between MDD patients and HC participants, we use different methods to learn the BN structures from the test validation dataset established in Section \ref{sec: real_world}.
Based on the learned BN structures, we perform a two-sample proportion test for each published overlapping connection of MDD in Figure \ref{fig: connection}, and report the $p$-values in Table \ref{tab: Two-sample}.
The smaller $p$ value indicates that NOTEARS-PFL recognizes the connection more easily in MDD group than in HC group.
We also included test results for NOTEARS-SIG, GES, PC, and GS for comparison.
In order to ensure the consistency of measurement standards, we control these methods to learn the same sparsity network structure.

\begin{table}
\centering
\caption{Two-sample proportion tests for published  overlapped connections.}
\setlength{\tabcolsep}{3pt}
\resizebox{0.7\linewidth}{!}{
\begin{tabular}{@{}cccccc@{}}
\toprule
Connection       & NOTEARS-PFL     & NOTEARS-SIG & GES    & PC     & GC     \\ \midrule
INS.R$\rightarrow$INS.L   & \textbf{0.0002} & 0.0016 & 0.0022 & 0.0512 & 0.0057 \\
INS.R$\rightarrow$ORBmid.R & \textbf{0.0012} & 0.0029  & 0.0055 & 0.0234 & 0.0133 \\
THA.L $\rightarrow$CAU.R & \textbf{0.0016} & 0.0040 & 0.0126 & 0.1636 & 0.0264 \\
PCG.L$\rightarrow$PCG.R & \textbf{0.0040} & 0.0091  & 0.0283 & 0.0984 & 0.0549 \\
INS.R $\rightarrow$ACG.R & \textbf{0.0049} & 0.0234 & 0.0133 & 0.2625 & 0.0289 \\
STG.R$\rightarrow$MTG.R & \textbf{0.0055} & 0.0264  & 0.0549 & 0.0721 & 0.0567 \\
MTG.L$\rightarrow$ORBmid.R & \textbf{0.0126} & 0.0283 & 0.0264 & 0.1556 & 0.1018 \\ \bottomrule
\end{tabular}
}
\label{tab: Two-sample}
\end{table}

\section{Conclusion}\label{sec:conclusion}

In this work, we focus on learning the heterogeneous brain functional connectivity in MDD patients using multi-site data efficiently and with the data sharing constraint.
To this end, we developed a toolbox called NOTEARS-PFL.
To achieve the model heterogeneity and computational efficiency, NOTEARS-PFL is formulated as minimizing a group fused lasso penalized score function with a continuous constraint for DAG. 
Considering the data-sharing barriers, the objective is solved in a federated learning framework using the ADMM, where original data is not exchanged during the optimization process. 
The algorithm is fast because the local update step has a closed-form expression, and the global update step can be solved using an efficient iterative proximal projection method. 
Extensive experiments on synthetic data and real-world multi-site rs-fMRI datasets with MDD demonstrate the  excellent performance of the proposed method. 
We highlight the usefulness of NOTERAS-PFL in analyzing multi-site rs-fMRI data and facilitating the study of MDD.

\bibliography{main}
\bibliographystyle{plainnat}

\appendix
\begin{appendix}
	\onecolumn
	\begin{center}
		{\huge {Supplementary Materials}}
	\end{center}

\section{Derivation of Federated Local Update Step}\label{sec:local_update}
To solve the subproblem as shown in equation \eqref{eq:W_k}, we first drop the subscript $t$ to lighten the notation, which leads to 
\begin{equation}
\underset{\boldsymbol{W}^{k}}{\min }f(\boldsymbol{W}^{k}):= \mathcal{L}(\boldsymbol{W}^{k},\mathbf{x}^{k})+\operatorname{tr}(\beta_{k}(\boldsymbol{W}^{k}-\boldsymbol{Z}^{k})^{\top})+\frac{\rho_{2}}{2}\left\|\boldsymbol{W}^{k}-\boldsymbol{Z}^{k}\right\|_{F}^{2}.
\label{eq:f(W_k)}
\end{equation}

Let $U^{k}=\frac{1}{n_{k}}(\mathbf{x}^{k})^{\top} \mathbf{x}^{k}$.
The first term of the equation \eqref{eq:f(W_k)} can be written as
\[
\begin{aligned} 
\mathcal{L}(\boldsymbol{W}^{k},\mathbf{x}^{k})&=\frac{1}{2 n_{k}}\left\|\mathbf{x}^{k}-\mathbf{x}^{k} \boldsymbol{W}^{k}\right\|_{F}^{2}, \\ &=\frac{1}{2 n_{k}} \operatorname{tr}((\mathbf{x}^{k}-\mathbf{x}^{k} \boldsymbol{W}^{k})^{\top}(\mathbf{x}^{k}-\mathbf{x}^{k} \boldsymbol{W}^{k})),\\ &=
\frac{1}{2 n_{k}} \operatorname{tr}((\mathbf{x}^{k})^{\top} \mathbf{x}^{k}-(\boldsymbol{W}^{k})^{\top} (\mathbf{x}^{k})^{\top} \mathbf{x}^{k}-(\mathbf{x}^{k})^{\top} \mathbf{x}^{k} \boldsymbol{W}^{k}+(\boldsymbol{W}^{k})^{\top} (\mathbf{x}^{k})^{\top} \mathbf{x}^{k} \boldsymbol{W}^{k}), \\ &
=\frac{1}{2} \operatorname{tr}\left(U^{k}-(\boldsymbol{W}^{k})^{\top} U^{k}-U^{k} \boldsymbol{W}^{k}+(\boldsymbol{W}^{k})^{\top} U^{k} \boldsymbol{W}^{k}\right), \\ &
=\frac{1}{2} \operatorname{tr}\left(U^{k}-2 (\boldsymbol{W}^{k})^{\top} U^{k}+(\boldsymbol{W}^{k})^{\top} U^{k} \boldsymbol{W}^{k}\right).
\end{aligned}
\]

Similarly, the third term of equation \eqref{eq:f(W_k)} can be written as 
\[
\begin{aligned}
\frac{\rho_{2}}{2}\|\boldsymbol{W}^{k}-\boldsymbol{Z}^{k}\|_{F}^{2}&=\frac{\rho_{2}}{2} \operatorname{tr}\left(\left(\boldsymbol{W}^{k}-\boldsymbol{Z}^{k}\right)^{\top}\left(\boldsymbol{W}^{k}-\boldsymbol{Z}^{k}\right)\right),\\ 
&=\frac{\rho_{2}}{2} \operatorname{tr}\left((\boldsymbol{W}^{k})^{\top} \boldsymbol{W}^{k}-2 (\boldsymbol{W}^{k})^{\top} \boldsymbol{Z}^{k}+(\boldsymbol{Z}^{k})^{\top} \boldsymbol{Z}^{k}\right). 
\end{aligned}
\]

Therefore, we have
\[
\begin{aligned}
f(\boldsymbol{W}^{k})&=\frac{1}{2} \operatorname{tr}(U^{k}-2 (\boldsymbol{W}^{k})^{\top} U^{k}+(\boldsymbol{W}^{k})^{\top} U^{k} \boldsymbol{W}^{k})+\operatorname{tr}(\beta_{k} (\boldsymbol{W}^{k})^{\top}-\beta_{k} (\boldsymbol{Z}^{k})^{\top})+\frac{\rho_{2}}{2} \operatorname{tr}((\boldsymbol{W}^{k})^{\top} \boldsymbol{W}^{k}\\
&\quad-2 (\boldsymbol{W}^{k})^{\top} \boldsymbol{Z}^{k}+(\boldsymbol{Z}^{k})^{\top} \boldsymbol{Z}^{k}), \\
&=\frac{1}{2} \operatorname{tr}(-2 (\boldsymbol{W}^{k})^{\top} U^{k}+(\boldsymbol{W}^{k})^{\top} U^{k} \boldsymbol{W}^{k})+\operatorname{tr}(\beta_{k} (\boldsymbol{W}^{k})^{\top}) \frac{\rho_{2}}{2} \operatorname{tr}((\boldsymbol{W}^{k})^{\top} \boldsymbol{W}^{k}-2 (\boldsymbol{W}^{k})^{\top} \boldsymbol{Z}^{k}+c,
\end{aligned}
\]
and its derivative is given by
\[
\begin{aligned}
\nabla_{\boldsymbol{W}^{k}} f\left(\boldsymbol{W}^{k}\right)&=\frac{1}{2} (-2 U^{k}+((U^{k})^{\top}+U^{k})\boldsymbol{W}^{k})+\beta_{k}+\frac{\rho_{2}}{2}(2 \boldsymbol{W}^{k}-2 \boldsymbol{Z}^{k}),\\
&=-U^{k}+U^{k} \boldsymbol{W}^{k}+\beta_{k}+\rho_{2} \boldsymbol{W}^{k}-\rho_{2}\boldsymbol{Z}^{k},\\
&= (U^{k}+\rho_{2} I) \boldsymbol{W}^{k}-U^{k}+\beta_{k}-\rho_{2} \boldsymbol{Z}^{k}.
\end{aligned}
\]
Assuming $U^{k}+\rho_{2}I$ is invertible, solving $\nabla_{\boldsymbol{W}^{k}} f\left(\boldsymbol{W}^{k}\right)=0$ yields the solution
\[
\boldsymbol{W}^{k}=(U^{k}+\rho_{2} I)^{-1}(\rho_{2} \boldsymbol{Z}^{k}-\beta_{k}+U^{k}).
\]

\section{Details on the Federated Global Update Step}\label{sec:global_update}
We provide details of the DIPA algorithm in the federated global update step.
For convenience, we introduce the following procedures and discuss how to transform $K$ matrices into one matrix, so that we can re-write equation \eqref{eq: proximal_operator} into vector form.

\textbf{Transform:} Given $K$ matrices $\tilde{\boldsymbol{M}}=\{{M^{1}, \ldots, M^{K}}\}$.
For each $k=1, \ldots, K$, $M^{k} \in\mathbb{R}^{d \times d}$.
Transform $M^{k}$ into vector form, denote as $m^{k}=\left(M_{i, j}^{k} \mid \text { for } i, j=1, \ldots, d\right)^{\top}$.
Then, construct a matrix $\Bar{\boldsymbol{M}}=\left(m^1, \ldots, m^{K}\right)^{\top}\in\mathbb{R}^{K \times d^2}$.

\textbf{Inverse-Transform:} The inverse-transform reverses the steps of the transform, which to transform the matrix $\Bar{\boldsymbol{M}}\in\mathbb{R}^{K \times d^2}$ to  $K$ matrices $\tilde{\boldsymbol{M}}=\{{M^{1}, \ldots, M^{K}}\}$.

The core of DIPA algorithm is to compute the separate proximal  operators for $\mathcal{R}_{1}$ and $\mathcal{R}_{2}$.
And then the iterative projection can be used to find a feasible point.
The proximal operator for the  $\ell_1$ term $\operatorname{prox}_{\mathcal{R}_{1}}\left(\Bar{\boldsymbol{U}}\right)$ is given by the soft-thresholding operator \citep{tibshirani1996regression}.

\begin{equation}
\begin{aligned}
\operatorname{prox}_{\mathcal{R}_{1}}\left(\Bar{\boldsymbol{U}} ; \lambda_1\right) &=\underset{\Bar{\boldsymbol{Z}}}{\arg \min } \frac{1}{2}\left\|\Bar{\boldsymbol{Z}}-\Bar{\boldsymbol{U}}\right\|_F^2+\lambda_1\|\Bar{\boldsymbol{Z}}\|_1,\\
&=\operatorname{sign}(\Bar{\boldsymbol{U}}) \odot \max \left(|\Bar{\boldsymbol{U}}|-\lambda_1, \mathbf{0}\right).
\end{aligned}
\label{eq:lasso}
\end{equation}
where the max and sign functions act in an element-wise manner and $\odot$ denotes element-wise multiplication.

The calculation of group-fused lasso proximal operator $\operatorname{prox}_{ \mathcal{R}_{2}}\left(\Bar{\boldsymbol{U}}\right)$ is more complex, and there is no obvious closed-form solution.
We tackle this through a block-coordinate descent approach \citep{bleakley2011group}.
Then, the proximal operator for the group smoothing aspect of the regularizer can be written as:

\begin{equation}
\operatorname{prox}_{\mathcal{R}_{2}}\left(\Bar{\boldsymbol{U}} ; \lambda_2\right) =\underset{\Bar{\boldsymbol{Z}}}{\arg \min } \frac{1}{2}\left\|\Bar{\boldsymbol{Z}}-\Bar{\boldsymbol{U}}\right\|_F^2+\lambda_2\|\boldsymbol{D}\Bar{\boldsymbol{Z}}\|_{2,1}.
\label{eq:group_fused}
\end{equation}

Let $\boldsymbol{A}=\boldsymbol{D}\Bar{\boldsymbol{Z}}$ and construct $\Bar{\boldsymbol{Z}}$ as a sum of differences via $Z_{k, \bullet}=\boldsymbol{a}+\sum_{i=1}^{K-1}\boldsymbol{A}_{i, \bullet},\left(\text { where } \boldsymbol{a}=Z_{1, \bullet}\right)$.
Then we can interpret the proximal operator as a group lasso probem \citep{bleakley2011group}.

\begin{equation}
\boldsymbol{A}:=\underset{\boldsymbol{A}}{\arg \min } \frac{1}{2}\|\Bar{\boldsymbol{U}}_{cen}-\Bar{\boldsymbol{R}}_{cen} \boldsymbol{A}\|_F^2+\lambda_2\|\boldsymbol{A}\|_{2,1},
\label{eq:A_hat}
\end{equation}
where $\boldsymbol{X}_{cen}$ denotes a column centered matrix and $\boldsymbol{R} \in \mathbb{R}^{K \times(K-1)}$ is a matrix with entries $R_{i,j}=1$ for $i>j$ and $0$ otherwise.
The problem above can be solved through a block-coordinate descent strategy, sequentially updating the solution for each block $\boldsymbol{A}_{k, \bullet}$ for $k=1, \ldots, K-1$.
We can then construct a solution for  $\Bar{\boldsymbol{Z}}$ by summing the differences.
And the optimal value for $\boldsymbol{a}$ is given by $\boldsymbol{a}=\mathbf{1}_{1, T}(\boldsymbol{U}-\boldsymbol{R} \boldsymbol{A})$. 
Correspondingly, the proximal operator for $\mathcal{R}_{2}$ is 
\begin{equation}
\operatorname{prox}_{\mathcal{R}_{2}}\left(\Bar{\boldsymbol{U}} ; \lambda_2\right)
=\left(\boldsymbol{a}^{\top},\left(\boldsymbol{a}+\boldsymbol{A}_{1, \bullet}\right)^{\top}, \ldots,\left(\boldsymbol{a}+\sum_{i=1}^{K-1} \boldsymbol{A}_{i, \bullet}\right)^{\top}\right)^{\top}.
\label{eq:R_2}
\end{equation}

\section{Details of the Synthetic Data Generating}\label{sec:synthetic_data}
We followed the similar synthetic data generate procedure as in \citep{liu2022learning}.
The key is how to create similar but different graphs by applying perturbations.

\textbf{Perturbation level:} Denote the adjacency matrix of $G_{truth}$ as $A_{t}=(a^{t}_{ij})\in\mathbb{R}^{d\times d}$, where $a^{t}_{ij}=1$ if there is an edge from node $i$ to node $j$, and $a^{t}_{ij}=0$ otherwise, $i,j=1,\ldots,d$.
Then the generated BN, denoted by $G_{gen}$, is constructed in the following way. 
Let $A_{g}$ be the adjacency matrix of $G_{gen}$, and define the perturbation level as the ratio of the number of changed edges in $A_{g}$ to the number of all possible edges. 
For example, if the perturbation level is $10\%$, we first set $A_{g}=A_{t}$, and then randomly select $10\%$ of the off-diagonal elements $a^{t}_{ij}(i\neq j)$ in $A_{t}$.
For each selected element $a^{t}_{ij}$, if $a^{t}_{ij}=0$, then we set $a^{g}_{ij}=1$ (adding edge); if $a^{t}_{ij}=1$, then with probability $1/2$ we set $a^{g}_{ij}=0$ (deleting edge), otherwise let $a^{g}_{ji}=1$ (reversing edge).
We generate $K$ graphs $\tilde{G}= \{G^{1}, \ldots, G^{K}\}$ by such a perturbation scheme.

\textbf{Evaluation metric:} Let $\{G_{output}^{1}, \ldots, G_{output}^{K}\}$ represent the $K$ BN structures obtained from NOTEARS-PFL, NOTEARS-SIG, NOTEARS-AVG, NOTEARS-ADMM.
The four basic statistics in the evaluation metrics are defined as follows: true positives (TP), the number of edges that are both in $G_{output}^{1}$ and $G^{1}$; false positives (FP), the number of edges that are present in $G_{output}^{1}$ but not in $G^{1}$; false negatives (FN), the number of edges that are present in $G^{1}$ but not in $G_{output}^{1}$; and true negatives (TN), the number of vertex pairs that are neither edges in $G^{1}$ nor in $G_{output}^{1}$. 

\newpage

\section{Supplementary Tables}\label{sec:supplementary_tables}

\begin{table}[h]
\centering
\caption{Data summary of the datasets used in this study.}
\setlength{\tabcolsep}{3pt}
\resizebox{0.9\linewidth}{!}{
\begin{tabular}{@{}cccccc@{}}
\toprule
\multicolumn{1}{l}{} & \begin{tabular}[c]{@{}c@{}}Center of Innovation in Hiroshima\\  (COI)\end{tabular} & \begin{tabular}[c]{@{}c@{}}Kyoto University \\ (KUT)\end{tabular} & \begin{tabular}[c]{@{}c@{}}University of Tokyo \\ (UTO)\end{tabular} & \begin{tabular}[c]{@{}c@{}}Hiroshima Kajikawa Hospital \\ (HKH)\end{tabular} & \begin{tabular}[c]{@{}c@{}}Hiroshima University Hospital \\ (HUH)\end{tabular} \\ \midrule
Total Subject        & 194  & 175  & 158 & 62  & 124 \\
MDD Subject          & 70   & 16   & 62 & 33 & 57   \\
HC Subject           & 124  & 159  & 96 & 29 & 67    \\
MDD Percentage       & 36\%  & 9\% & 39\%  & 53\%  & 46\% \\
rs-fMRI Frames       & 240  & 240  & 240 & 107 & 143 \\ \bottomrule
\end{tabular}
}
\label{tab: summary}
\end{table}

\begin{table}[h]
\centering
\caption{Demographic characteristics of participants in multi-site datasets.}
%\label{table}
\setlength{\tabcolsep}{3pt}
\resizebox{0.4\linewidth}{!}{
\begin{tabular}{@{}ccccll@{}}
\toprule
    & Site & Number & Male/Female & \multicolumn{1}{c}{Age} & \multicolumn{1}{c}{BDI} \\ \midrule
MDD & COI  & 70     & 31/39       & 45.0 ± 12.5             & 26.2 ± 9.9              \\
    & KUT  & 16     & 10 6        & 33.79±10.82             & 27.7 ± 10.1             \\
    & UTO  & 62     & 36/26       & 38.74 ±11.62            & 21.5±11.3               \\
    & HKH  & 33     & 20/13       & 44.8±11.5               & 28.5±8.7                \\
    & HUH  & 57     & 32/25       & 43.3±12.2               & 30.9±9.0                \\
HC  & COI  & 124    & 46/78       & 51.9 ± 13.4             & 8.2 ± 6.3               \\
    & KUT  & 159    & 93/66       & 36.51± 13.59            & 6.0 ± 5.4               \\
    & UTO  & 96     & 33/63       & 46.65±15.54             & 6.6 ± 6.5               \\
    & HKH  & 29     & 12 17       & 45.4±9.5                & 5.1±4.6                 \\
    & HUH  & 67     & 30/37       & 45.6±9.4                & 5.6±4.3                 \\ \bottomrule
\end{tabular}
}
\label{tab:demographic}
\end{table}

\begin{table}[h]
\centering
\caption{Imaging protocols for rs-fMRI and structural MRI in multi-site datasets. }
\setlength{\tabcolsep}{3pt}
\resizebox{0.75\linewidth}{!}{
\begin{tabular}{@{}clccccc@{}}
\toprule
               & \multicolumn{1}{c}{}  & COI & KUT  & UTO  & HKH & HUH  \\ \midrule
rs-fMRI        & MRI scanner & \begin{tabular}[c]{@{}c@{}}Siemens \\ Verio\end{tabular}        & \begin{tabular}[c]{@{}c@{}}Siemens \\ TimTrio\end{tabular}  & \begin{tabular}[c]{@{}c@{}}GE\\  MR750w\end{tabular}  & \begin{tabular}[c]{@{}c@{}}Siemens\\  Spectra\end{tabular}   & \begin{tabular}[c]{@{}c@{}}GE\\  Sigma HDxt\end{tabular}             \\& Magnetic field strength & 3.0T  & 3.0T  & 3.0T   & 3.0T & 3.0T \\& Number of channels per coil   & 12 & 32  & 24  & 12  & 8  \\ & FoV (mm)  & 212 & 212 & 212 & 192 & 256\\ & Matrix     & 64 x 64  & 64 x 64 & 64 x 64 & 64 x 64 & 64 x 64 \\
 & Number of slices & 40    & 40 & 40  & 38 & 32 \\
 & Number of volumes & 240  & 240  & 240 & 107  & 143  \\
 & In-plane resolution (mm) & 3.3 x 3.3 & 3.3125 X 3.3125 & 3.3 x 3.3 & 3.0 x 3.0 & 4.0 x 4.0\\ & Slice thickness (mm) & 3.2 & 3.2 & 3.2 & 3 & 4 \\
& Slice gap (mm) & 0.8 & 0.8 & 0.8   & 0   & 0  \\
& TR (ms)  & 2500  & 2500  & 2500   & 2700 & 2000\\
 & TE (ms)   & 30  & 30& 30  & 31 & 27  \\ & Flip angel (deg)  & 80 & 80 & 80   & 90& 90\\
& Slice acquisition order  & Ascending  & Ascending & Ascending  & Ascending & Ascending       \\& \begin{tabular}[c]{@{}l@{}}Total scan\\  time\end{tabular} & \begin{tabular}[c]{@{}c@{}}10 min + 10s\\  (dummy)\end{tabular} & \begin{tabular}[c]{@{}c@{}}10 min + 10s \\ (dummy)\end{tabular} & \begin{tabular}[c]{@{}c@{}}10 min + 10s\\  (dummy)\end{tabular} & \begin{tabular}[c]{@{}c@{}}4 min. 46s. + 14s \\ (dummy)\end{tabular} & \begin{tabular}[c]{@{}c@{}}4 min. 46s. + 14s \\ (dummy)\end{tabular} \\& Eye closed/fixate     & Fixate& Fixate& Fixate & Fixate                & Fixate  \\ Structural MRI & FoV (mm) & 256   & 225x 240  & 240  & 256& 256   \\
& Matrix   & 256 x 256 & 240 x 256    & 256 x 256& 256 x 256    & 256 x 256 \\ & Voxel size mm3  & 1 x 1x 1 & 0.9375 x 0.9375 x 1  & 1 x 1x 1.2  & 1 x 1x 1  & 1 x 1x 1 \\ & TR (ms) & 2300  & 2000 & 7.7  & 1900  & 6812  \\ & TE (ms) & 2.98 & 3.4 & 3.1 & 2.38   & 1986  \\
& TI (ms)  & 900 & 990 & 400 & 900 & 450 \\ & Flip angel (deg) & 9 & 8 & 11 & 10 & 20       \\ \bottomrule
\end{tabular}
}
\label{tab: imaging_protocol}
\end{table}

\begin{table}
\centering
\caption{Description of ROIs in AAL template and their Montreal Neurological Institute (MNI) coordinates.}
%\label{table}
\setlength{\tabcolsep}{3pt}
\resizebox{0.75\linewidth}{!}{
\begin{tabular}{@{}clccccc@{}}
\toprule
Index & \multicolumn{1}{c}{Region name}          & Abbreviation & Lobe                      & x      & y      & z      \\ \midrule
1     & \textbf{Precentral gyrus}                        & PreCG.L      & Frontal                   & -38.65 & -5.68  & 50.94  \\
2     & \textbf{Precentral gyrus}                        & PreCG.R      & Frontal                   & 41.37  & -8.21  & 52.09  \\
3     & \textbf{Superior frontal gyrus, dorsolateral}     & SFGdor.L     & Frontal                   & -18.45 & 34.81  & 42.20  \\
4     & \textbf{Superior frontal gyrus, dorsolateral}     & SFGdor.R     & Frontal                   & 21.90  & 31.12  & 43.82  \\
5     & \textbf{Superior frontal gyrus, orbital}          & ORBsup.L     & Frontal                   & -16.56 & 47.32  & -13.31 \\
6     & \textbf{Superior frontal gyrus, orbital}          & ORBsup.R     & Frontal                   & 18.49  & 48.10  & -14.02 \\
7     & \textbf{Middle frontal gyrus}                    & MFG.L        & Frontal                   & -33.43 & 32.73  & 35.46  \\
8     & \textbf{Middle frontal gyrus}                     & MFG.R        & Frontal                   & 37.59  & 33.06  & 34.04  \\
9     & \textbf{Middle frontal gyrus, orbital}            & ORBmid.L     & Frontal                   & -30.65 & 50.43  & -9.62  \\
10    & \textbf{Middle frontal gyrus, orbital}            & ORBmid.R     & Frontal                   & 33.18  & 52.59  & -10.73 \\
11    & \textbf{Inferior frontal gyrus, opercular}        & IFGoperc.L   & Frontal                   & -48.43 & 12.73  & 19.02  \\
12    & \textbf{Inferior frontal gyrus, opercular}        & IFGoperc.R   & Frontal                   & 50.20  & 14.98  & 21.41  \\
13    & \textbf{Inferior frontal gyrus, triangular}       & IFGtriang.L  & Frontal                   & -45.58 & 29.91  & 13.99  \\
14    & \textbf{Inferior frontal gyrus, triangular}       & IFGtriang.R  & Frontal                   & 50.33  & 30.16  & 14.17  \\
15    & \textbf{Inferior frontal gyrus, orbital}          & ORBinf.L     & Frontal                   & -35.98 & 30.71  & -12.11 \\
16    & \textbf{Inferior frontal gyrus, orbital}          & ORBinf.R     & Frontal                   & 41.22  & 32.23  & -11.91 \\
17    & Rolandic operculum                       & ROL.L        & Frontal                   & -47.16 & -8.48  & 13.95  \\
18    & Rolandic operculum                       & ROL.R        & Frontal                   & 52.65  & -6.25  & 14.63  \\
19    & Supplementary motor area                 & SMA.L        & Frontal                   & -5.32  & 4.85   & 61.38  \\
20    & Supplementary motor area                 & SMA.R        & Frontal                   & 8.62   & 0.17   & 61.85  \\
21    & Olfactory cortex                         & OLF.L        & Frontal                   & -8.06  & 15.05  & -11.46 \\
22    & Olfactory cortex                         & OLF.R        & Frontal                   & 10.43  & 15.91  & -11.26 \\
23    & \textbf{Superior frontal gyrus, medial}           & SFGmed.L     & Frontal                   & -4.80  & 49.17  & 30.89  \\
24    & \textbf{Superior frontal gyrus, medial}           & SFGmed.R     & Frontal                   & 9.10   & 50.84  & 30.22  \\
25    & \textbf{Superior frontal gyrus, medial orbital}   & ORBsupmed.L  & Frontal                   & -5.17  & 54.06  & -7.40  \\
26    & \textbf{Superior frontal gyrus, medial orbital}   & ORBsupmed.R  & Frontal                   & 8.16   & 51.67  & -7.13  \\
27    & Gyrus rectus                             & REC.L        & Frontal                   & -5.08  & 37.07  & -18.14 \\
28    & Gyrus rectus                             & REC.R        & Frontal                   & 8.35   & 35.64  & -18.04 \\
29    & \textbf{Insula}                                   & INS.L        & Insula and Cingulate Gyri & -35.13 & 6.65   & 3.44   \\
30    & \textbf{Insula}                                   & INS.R        & Insula and Cingulate Gyri & 39.02  & 6.25   & 2.08   \\
31    & \textbf{Cingulate gyrus, anterior part}           & ACG.L        & Insula and Cingulate Gyri & -4.04  & 35.40  & 13.95  \\
32    & \textbf{Cingulate gyrus, anterior part}           & ACG.R        & Insula and Cingulate Gyri & 8.46   & 37.01  & 15.84  \\
33    & \textbf{Cingulate gyrus, mid part}               & DCG.L        & Insula and Cingulate Gyri & -5.48  & -14.92 & 41.57  \\
34    & \textbf{Cingulate gyrus, mid part}                & DCG.R        & Insula and Cingulate Gyri & 8.02   & -8.83  & 39.79  \\
35    & \textbf{Cingulate gyurs, posterior part}          & PCG.L        & Insula and Cingulate Gyri & -4.85  & -42.92 & 24.67  \\
36    & \textbf{Cingulate gyurs, posterior part}          & PCG.R        & Insula and Cingulate Gyri & 7.44   & -41.81 & 21.87  \\
37    & \textbf{Hippocampus}                              & HIP.L        & Temporal                  & -25.03 & -20.74 & -10.13 \\
38    & \textbf{Hippocampus}                              & HIP.R        & Temporal                  & 29.23  & -19.78 & -10.33 \\
39    & \textbf{Parahippocampus}                          & PHG.L        & Temporal                  & -21.17 & -15.95 & -20.70 \\
40    & \textbf{Parahippocampus}                          & PHG.R        & Temporal                  & 25.38  & -15.15 & -20.47 \\
41    & \textbf{Amygdala}                                 & AMYG.L       & Temporal                  & -23.27 & -0.67  & -17.14 \\
42    & \textbf{Amygdala}                                 & AMYG.R       & Temporal                  & 27.32  & 0.64   & -17.50 \\
43    & Calcarine fissure and surrounding cortex & CAL.L        & Occipital                 & -7.14  & -78.67 & 6.44   \\
44    & Calcarine fissure and surrounding cortex & CAL.R        & Occipital                 & 15.99  & -73.15 & 9.40   \\
45    & \textbf{Cuneus}                                   & CUN.L        & Occipital                 & -5.93  & -80.13 & 27.22  \\
46    & \textbf{Cuneus}                                   & CUN.R        & Occipital                 & 13.51  & -79.36 & 28.23  \\
47    & Lingual gyrus                            & LING.L       & Occipital                 & -14.62 & -67.56 & -4.63  \\
48    & Lingual gyrus                            & LING.R       & Occipital                 & 16.29  & -66.93 & -3.87  \\
49    & Superior occipital lobe                  & SOG.L        & Occipital                 & -16.54 & -84.26 & 28.17  \\
50    & Superior occipital lobe                  & SOG.R        & Occipital                 & 24.29  & -80.85 & 30.59  \\
51    & Middle occipital lobe                    & MOG.L        & Occipital                 & -32.39 & -80.73 & 16.11  \\
52    & Middle occipital lobe                    & MOG.R        & Occipital                 & 37.39  & -79.70 & 19.42  \\
53    & Inferior occipital lobe                  & IOG.L        & Occipital                 & -36.36 & -78.29 & -7.84  \\
54    & Inferior occipital lobe                  & IOG.R        & Occipital                 & 38.16  & -81.99 & -7.61  \\
55    & \textbf{Fusiform gyrus}                           & FFG.L        & Temporal                  & -31.16 & -40.30 & -20.23 \\
56    & \textbf{Fusiform gyrus}                           & FFG.R        & Temporal                  & 33.97  & -39.10 & -20.18 \\
57    & \textbf{Postcentral gyrus}                        & PoCG.L       & Pariental                 & -42.46 & -22.63 & 48.92  \\
58    & \textbf{Postcentral gyrus}                        & PoCG.R       & Pariental                 & 41.43  & -25.49 & 52.55  \\
59    & Superior pariental gyrus                 & SPG.L        & Pariental                 & -23.45 & -59.56 & 58.96  \\
60    & Superior pariental gyrus                 & SPG.R        & Pariental                 & 26.11  & -59.18 & 62.06  \\
61    & Inferior pariental gyrus                 & IPL.L        & Pariental                 & -42.80 & -45.82 & 46.74  \\
62    & Inferior pariental gyrus                 & IPL.R        & Pariental                 & 46.46  & -46.29 & 49.54  \\
63    & Supramarginal gyrus                      & SMG.L        & Pariental                 & -55.79 & -33.64 & 30.45  \\
\bottomrule
\end{tabular}
}
\label{tab: roi}
\end{table}

\begin{table*}
\centering
\caption{Continue table: Description of ROIs in AAL template and their Montreal Neurological Institute (MNI) coordinates.}
%\label{table}
\setlength{\tabcolsep}{3pt}
\resizebox{0.75\linewidth}{!}{
\begin{tabular}{@{}clccccc@{}}
\toprule
Index & \multicolumn{1}{c}{Region name}        & Abbreviation & Lobe            & x      & y      & z      \\ \midrule
64    & Supramarginal gyrus                      & SMG.R        & Pariental                 & 57.61  & -31.50 & 34.48  \\
65    & Angular gyrus                            & ANG.L        & Pariental                 & -44.14 & -60.82 & 35.59  \\
66    & Angular gyrus                            & ANG.R        & Pariental                 & 45.51  & -59.98 & 38.63  \\
67    & \textbf{Precuneus}                                & PCUN.L       & Pariental                 & -7.24  & -56.07 & 48.01  \\
68    & \textbf{Precuneus}                                & PCUN.R       & Pariental                 & 9.98   & -56.05 & 43.77  \\
69    & Paracentral lobule                       & PCL.L        & Frontal                   & -7.63  & -25.36 & 70.07  \\
70    & Paracentral lobule                       & PCL.R        & Frontal                   & 7.48   & -31.59 & 68.09  \\
71    & \textbf{Caudate}                                  & CAU.L        & Subcortial region         & -11.46 & 11.00  & 9.24   \\
72    & \textbf{Caudate}                                  & CAU.R        & Subcortial region         & 14.84  & 12.07  & 9.42   \\
73    & Putamen                                  & PUT.L        & Subcortial region         & -23.91 & 3.86   & 2.40   \\ 
74    & Putamen                                  & PUT.R        & Subcortial region         & 27.78  & 4.91   & 2.46 \\
75    & Pallidum                                 & PAL.L        & Subcortial region         & -17.75 & -0.03  & 0.21   \\
76    & Pallidum                                 & PAL.R        & Subcortial region         & 21.20  & 0.18   & 0.23   \\
77    & \textbf{Thalamus}                                 & THA.L        & Subcortial region         & -10.85 & -17.56 & 7.98 \\
78    & \textbf{Thalamus}                               & THA.R        & Temporal        & 13.00  & -17.55 & 8.09   \\
79    & \textbf{Heschl gyrus}                           & HES.L        & Temporal        & -41.99 & -18.88 & 9.98   \\
80    & \textbf{Heschl gyrus}                           & HES.R        & Temporal        & 45.86  & -17.15 & 10.41  \\
81    & \textbf{Superior temporal gyrus}                & STG.L        & Temporal        & -53.16 & -20.68 & 7.13   \\
82    & \textbf{Superior temporal gyrus}               & STG.R        & Temporal        & 58.15  & -21.78 & 6.80   \\
83    & Temporal pole: superior temporal gyrus & TPOsup.L     & Temporal        & -39.88 & 15.14  & -20.18 \\
84    & Temporal pole: superior temporal gyrus & TPOsup.R     & Temporal        & 48.25  & 14.75  & -16.86 \\
85    & \textbf{Middle temporal gyrus}                  & MTG.L        & Temporal        & -55.52 & -33.80 & -2.20  \\
86    & \textbf{Middle temporal gyrus}                  & MTG.R        & Temporal        & 57.47  & -37.23 & -1.47  \\
87    & \textbf{Temporal pole: middle temporal gyrus}   & TPOmid.L     & Temporal        & -36.32 & 14.59  & -34.08 \\
88    & \textbf{Temporal pole: middle temporal gyrus}   & TPOmid.R     & Temporal        & 44.22  & 14.55  & -32.23 \\
89    & \textbf{Inferior temporal gyrus}                & ITG.L        & Temporal        & -49.77 & -28.05 & -23.17 \\
90    & \textbf{Inferior temporal gyrus}                & ITG.R        & Temporal        & 53.69  & -31.07 & -22.32 \\
91    & Cerebellum crus 1                      & CRBLCrus1.L  & Posterior Fossa & -36.07 & -66.72 & -28.93 \\
92    & Cerebellum crus 1                      & CRBLCrus1.R  & Posterior Fossa & 37.46  & -67.14 & -29.55 \\
93    & Cerebellum crus 2                      & CRBLCrus2.L  & Posterior Fossa & -28.64 & -73.26 & -38.2  \\
94    & Cerebellum crus 2                      & CRBLCrus2.R  & Posterior Fossa & 32.06  & -69.02 & -39.95 \\
95    & Cerebellum 3                           & CRBL3.L      & Posterior Fossa & -8.8   & -37.22 & -18.58 \\
96    & Cerebellum 3                           & CRBL3.R      & Posterior Fossa & 12.32  & -34.47 & -19.39 \\
97    & Cerebellum 4                           & CRBL45.L     & Posterior Fossa & -15    & -43.49 & -16.93 \\
98    & Cerebellum 4                           & CRBL45.R     & Posterior Fossa & 17.2   & -42.86 & -18.15 \\
99    & Cerebellum 6                           & CRBL6.L      & Posterior Fossa & -23.24 & -59.1  & -22.13 \\
100   & Cerebellum 6                           & CRBL6.R      & Posterior Fossa & 24.69  & -58.32 & -23.65 \\
101   & Cerebellum 7                           & CRBL7b.L     & Posterior Fossa & -32.36 & -59.82 & -45.45 \\
102   & Cerebellum 7                           & CRBL7b.R     & Posterior Fossa & 33.14  & -63.18 & -48.46 \\
103   & Cerebellum 8                           & CRBL8.L      & Posterior Fossa & -25.75 & -54.52 & -47.68 \\
104   & Cerebellum 8                           & CRBL8.R      & Posterior Fossa & 25.06  & -56.34 & -49.47 \\
105   & Cerebellum 9                           & CRBL9.L      & Posterior Fossa & -10.95 & -48.95 & -45.9  \\
106   & Cerebellum 9                           & CRBL9.R      & Posterior Fossa & 9.46   & -49.5  & -46.33 \\
107   & Cerebellum 10                          & CRBL10.L     & Posterior Fossa & -22.61 & -33.8  & -41.76 \\
108   & Cerebellum 10                          & CRBL10.R     & Posterior Fossa & 25.99  & -33.84 & -41.35 \\
109   & Vermis12                               & Vermis12     & Posterior Fossa & 0.76   & -38.79 & -20.05 \\
110   & Vermis3                                & Vermis3      & Posterior Fossa & 1.38   & -39.93 & -11.4  \\
111   & Vermis45                               & Vermis45     & Posterior Fossa & 1.22   & -52.36 & -6.11  \\
112   & Vermis6                                & Vermis6      & Posterior Fossa & 1.14   & -67.06 & -15.12 \\
113   & Vermis7                                & Vermis7      & Posterior Fossa & 1.15   & -71.93 & -25.14 \\
114   & Vermis8                                & Vermis8      & Posterior Fossa & 1.15   & -64.43 & -34.08 \\
115   & Vermis9                                & Vermis9      & Posterior Fossa & 0.86   & -54.87 & -34.9  \\
116   & Vermis10                               & Vermis10     & Posterior Fossa & 0.36   & -45.8  & -31.68 \\ \bottomrule
\end{tabular}
}
\label{tab: roi_continue}
\end{table*}

\begin{table*}
\centering
\caption{The overlapping connections between the directed networks of 5 sites of MDD patients and HC participants.}
%\label{table}
\setlength{\tabcolsep}{3pt}
\resizebox{0.75\linewidth}{!}{
\begin{tabular}{@{}lcccccc@{}}
\toprule
             & \textbf{} & \textbf{From} &                           & \textbf{} & \textbf{To} &                           \\ \midrule
             & Index     & ROI           & Lobe                      & Index     & ROI         & Lobe                      \\
\textbf{MDD} & 10        & ORBmid.R      & Frontal                   & 6         & ORBsup.R    & Frontal                   \\
             & 24        & SFGmed.R      & Frontal                   & 33        & DCG.L       & Insula and Cingulate Gyri \\
             & 30        & INS.R         & Insula and Cingulate Gyri & 10        & ORBmid.R    & Frontal                   \\
             & 30        & INS.R         & Insula and Cingulate Gyri & 29        & INS.L       & Insula and Cingulate Gyri \\
             & 30        & INS.R         & Insula and Cingulate Gyri & 32        & ACG.R       & Insula and Cingulate Gyri \\
             & 33        & DCG.L         & Insula and Cingulate Gyri & 3         & SFGdor.L    & Frontal                   \\
             & 35        & PCG.L         & Insula and Cingulate Gyri & 36        & PCG.R       & Insula and Cingulate Gyri \\
             & 46        & CUN.R         & Occipital                 & 45        & CUN.L       & Occipital                 \\
             & 56        & FFG.R         & Temporal                  & 11        & IFGoperc.L  & Frontal                   \\
             & 68        & PCUN.R        & Pariental                 & 46        & CUN.R       & Occipital                 \\
             & 68        & PCUN.R        & Pariental                 & 26        & ORBsupmed.R & Frontal                   \\
             & 71        & CAU.L         & Subcortial region         & 6         & ORBsup.R    & Frontal                   \\
             & 77        & THA.L         & Subcortial region         & 72        & CAU.R       & Subcortial region         \\
             & 82        & STG.R         & Temporal                  & 86        & MTG.R       & Temporal                  \\
             & 85        & MTG.L         & Temporal                  & 10        & ORBmid.R    & Frontal                   \\
             & 89        & ITG.L         & Temporal                  & 1         & PreCG.L     & Frontal                   \\
\textbf{HC}  & 1         & PreCG.L       & Frontal                   & 26        & ORBsupmed.R & Frontal                   \\
             & 8         & MFG.R         & Frontal                   & 9         & ORBmid.L    & Frontal                   \\
             & 23        & SFGmed.L      & Frontal                   & 24        & SFGmed.R    & Frontal                   \\
             & 31        & ACG.L         & Insula and Cingulate Gyri & 23        & SFGmed.L    & Frontal                   \\
             & 34        & DCG.R         & Insula and Cingulate Gyri & 67        & PCUN.L      & Pariental                 \\
             & 38        & HIP.R         & Temporal                  & 39        & PHG.L       & Temporal                  \\
             & 40        & PHG.R         & Temporal                  & 3         & SFGdor.L    & Frontal                   \\
             & 82        & STG.R         & Temporal                  & 8         & MFG.R       & Frontal                   \\
             & 86        & MTG.R         & Temporal                  & 89        & ITG.L       & Temporal                  \\ \bottomrule
\end{tabular}
}
\label{tab: overlapping_connections}
\end{table*}

\begin{table}
\centering
\caption{The site-specific connections between the directed networks of 5 sites of MDD patients.}
\setlength{\tabcolsep}{3pt}
\resizebox{0.65\linewidth}{!}{
\begin{tabular}{@{}ccccccc@{}}
\toprule
                &       & From        &                           &       & To          &                           \\ \midrule
                & Index & ROI         & Lobe                      & Index & ROI         & Lobe                      \\
\textbf{Site 1} & 1     & PreCG.L     & Frontal                   & 10    & ORBmid.R    & Frontal                   \\
                & 1     & PreCG.L     & Frontal                   & 67    & PCUN.L      & Pariental                 \\
                & 2     & PreCG.R     & Frontal                   & 67    & PCUN.L      & Pariental                 \\
                & 5     & ORBsup.L    & Frontal                   & 39    & PHG.L       & Temporal                  \\
                & 6     & ORBsup.R    & Frontal                   & 78    & THA.R       & Temporal                  \\
                & 7     & MFG.L       & Frontal                   & 71    & CAU.L       & Subcortial region         \\
                & 8     & MFG.R       & Frontal                   & 82    & STG.R       & Temporal                  \\
                & 11    & IFGoperc.L  & Frontal                   & 68    & PCUN.R      & Pariental                 \\
                & 13    & IFGtriang.L & Frontal                   & 67    & PCUN.L      & Pariental                 \\
                & 16    & ORBinf.R    & Frontal                   & 56    & FFG.R       & Temporal                  \\
                & 24    & SFGmed.R    & Frontal                   & 72    & CAU.R       & Subcortial region         \\
                & 26    & ORBsupmed.R & Frontal                   & 42    & AMYG.R      & Temporal                  \\
                & 29    & INS.L       & Insula and Cingulate Gyri & 8     & MFG.R       & Frontal                   \\
                & 42    & AMYG.R      & Temporal                  & 87    & TPOmid.L    & Temporal                  \\
                & 55    & FFG.L       & Temporal                  & 68    & PCUN.R      & Pariental                 \\
                & 77    & THA.L       & Subcortial region         & 31    & ACG.L       & Insula and Cingulate Gyri \\
                & 90    & ITG.R       & Temporal                  & 56    & FFG.R       & Temporal                  \\
\textbf{Site2}  & 3     & SFGdor.L    & Frontal                   & 7     & MFG.L       & Frontal                   \\
                & 4     & SFGdor.R    & Frontal                   & 1     & PreCG.L     & Frontal                   \\
                & 5     & ORBsup.L    & Frontal                   & 25    & ORBsupmed.L & Frontal                   \\
                & 7     & MFG.L       & Frontal                   & 10    & ORBmid.R    & Frontal                   \\
                & 15    & ORBinf.L    & Frontal                   & 11    & IFGoperc.L  & Frontal                   \\
                & 24    & SFGmed.R    & Frontal                   & 3     & SFGdor.L    & Frontal                   \\
                & 29    & INS.L       & Insula and Cingulate Gyri & 9     & ORBmid.L    & Frontal                   \\
                & 29    & INS.L       & Insula and Cingulate Gyri & 16    & ORBinf.R    & Frontal                   \\
                & 31    & ACG.L       & Insula and Cingulate Gyri & 5     & ORBsup.L    & Frontal                   \\
                & 77    & THA.L       & Subcortial region         & 13    & IFGtriang.L & Frontal                   \\
                & 81    & STG.L       & Temporal                  & 24    & SFGmed.R    & Frontal                   \\
                & 85    & MTG.L       & Temporal                  & 14    & IFGtriang.R & Frontal                   \\
                & 33    & DCG.L       & Insula and Cingulate Gyri & 35    & PCG.L       & Insula and Cingulate Gyri \\
                & 37    & HIP.L       & Temporal                  & 12    & IFGoperc.R  & Frontal                   \\
                & 89    & ITG.L       & Temporal                  & 42    & AMYG.R      & Temporal                  \\
                & 56    & FFG.R       & Temporal                  & 35    & PCG.L       & Insula and Cingulate Gyri \\
                & 86    & MTG.R       & Temporal                  & 5     & ORBsup.L    & Frontal                   \\
\textbf{Site3}  & 1     & PreCG.L     & Frontal                   & 14    & IFGtriang.R & Frontal                   \\
                & 2     & PreCG.R     & Frontal                   & 57    & PoCG.L      & Pariental                 \\
                & 5     & ORBsup.L    & Frontal                   & 26    & ORBsupmed.R & Frontal                   \\
                & 7     & MFG.L       & Frontal                   & 35    & PCG.L       & Insula and Cingulate Gyri \\
                & 11    & IFGoperc.L  & Frontal                   & 23    & SFGmed.L    & Frontal                   \\
                & 13    & IFGtriang.L & Frontal                   & 46    & CUN.R       & Occipital                 \\
                & 23    & SFGmed.L    & Frontal                   & 9     & ORBmid.L    & Frontal                   \\
                & 29    & INS.L       & Insula and Cingulate Gyri & 3     & SFGdor.L    & Frontal                   \\
                & 33    & DCG.L       & Insula and Cingulate Gyri & 42    & AMYG.R      & Temporal                  \\
                & 33    & DCG.L       & Insula and Cingulate Gyri & 45    & CUN.L       & Occipital                 \\
                & 35    & PCG.L       & Insula and Cingulate Gyri & 85    & MTG.L       & Temporal                  \\
                & 40    & PHG.R       & Temporal                  & 11    & IFGoperc.L  & Frontal                   \\
                & 39    & PHG.L       & Temporal                  & 6     & ORBsup.R    & Frontal                   \\
                & 72    & CAU.R       & Subcortial region         & 56    & FFG.R       & Temporal                  \\
                & 78    & THA.R       & Temporal                  & 2     & PreCG.R     & Frontal                   \\
                & 80    & HES.R       & Temporal                  & 82    & STG.R       & Temporal                  \\
\textbf{Site4}  & 2     & PreCG.R     & Frontal                   & 24    & SFGmed.R    & Frontal                   \\
                & 5     & ORBsup.L    & Frontal                   & 4     & SFGdor.R    & Frontal                   \\
                & 7     & MFG.L       & Frontal                   & 29    & INS.L       & Insula and Cingulate Gyri \\
                & 9     & ORBmid.L    & Frontal                   & 57    & PoCG.L      & Pariental                 \\
                & 12    & IFGoperc.R  & Frontal                   & 14    & IFGtriang.R & Frontal                   \\
                & 15    & ORBinf.L    & Frontal                   & 7     & MFG.L       & Frontal                   \\
                & 24    & SFGmed.R    & Frontal                   & 12    & IFGoperc.R  & Frontal                   \\
                & 30    & INS.R       & Insula and Cingulate Gyri & 23    & SFGmed.L    & Frontal                   \\
                & 31    & ACG.L       & Insula and Cingulate Gyri & 16    & ORBinf.R    & Frontal                   \\
                & 32    & ACG.R       & Insula and Cingulate Gyri & 68    & PCUN.R      & Pariental                 \\
                & 34    & DCG.R       & Insula and Cingulate Gyri & 5     & ORBsup.L    & Frontal                   \\
                & 37    & HIP.L       & Temporal                  & 29    & INS.L       & Insula and Cingulate Gyri \\
                & 40    & PHG.R       & Temporal                  & 16    & ORBinf.R    & Frontal                   \\
                & 41    & AMYG.L      & Temporal                  & 3     & SFGdor.L    & Frontal                   \\
                & 45    & CUN.L       & Occipital                 & 72    & CAU.R       & Subcortial region         \\
                & 56    & FFG.R       & Temporal                  & 7     & MFG.L       & Frontal                   \\
                & 68    & PCUN.R      & Pariental                 & 15    & ORBinf.L    & Frontal                   \\
                & 87    & TPOmid.L    & Temporal                  & 33    & DCG.L       & Insula and Cingulate Gyri \\
                & 90    & ITG.R       & Temporal                  & 86    & MTG.R       & Temporal                  \\ \bottomrule
\end{tabular}
}
\label{tab:site_variant }
\end{table}

\begin{table*}
\centering
\caption{Continue table: The site-specific connections between the directed networks of 5 sites of MDD patients.}
%\label{table}
\setlength{\tabcolsep}{3pt}
\resizebox{0.65\linewidth}{!}{
\begin{tabular}{@{}ccccccc@{}}
\toprule
                & \textbf{} & \textbf{From} &                           & \textbf{} & \textbf{To} &                           \\ \midrule
                & Index     & ROI           & Lobe                      & Index    & ROI         & Lobe                      \\
\textbf{Site 5}  & 1     & PreCG.L     & Frontal                   & 24    & SFGmed.R    & Frontal                   \\
                & 2     & PreCG.R     & Frontal                   & 46    & CUN.R       & Occipital                 \\
                & 7     & MFG.L       & Frontal                   & 2     & PreCG.R     & Frontal \\
                & 10        & ORBmid.R      & Frontal                   & 25        & ORBsupmed.L & Frontal                   \\
                & 13        & IFGtriang.L   & Frontal                   & 15        & ORBinf.L    & Frontal                   \\
                & 15        & ORBinf.L      & Frontal                   & 4         & SFGdor.R    & Frontal                   \\
                & 24        & SFGmed.R      & Frontal                   & 10        & ORBmid.R    & Frontal                   \\
                & 26        & ORBsupmed.R   & Frontal                   & 77        & THA.L       & Subcortial region         \\
                & 33        & DCG.L         & Insula and Cingulate Gyri & 31        & ACG.L       & Insula and Cingulate Gyri \\
                & 34        & DCG.R         & Insula and Cingulate Gyri & 38        & HIP.R       & Temporal                  \\
                & 36        & PCG.R         & Insula and Cingulate Gyri & 39        & PHG.L       & Temporal                  \\
                & 39        & PHG.L         & Temporal                  & 4         & SFGdor.R    & Frontal                   \\
                & 46        & CUN.R         & Occipital                 & 77        & THA.L       & Subcortial region         \\
                & 56        & FFG.R         & Temporal                  & 32        & ACG.R       & Insula and Cingulate Gyri \\
                & 67        & PCUN.L        & Pariental                 & 4         & SFGdor.R    & Frontal                   \\
                & 78        & THA.R         & Temporal                  & 39        & PHG.L       & Temporal                  \\
                & 86        & MTG.R         & Temporal                  & 16        & ORBinf.R    & Frontal                   \\
                & 89        & ITG.L         & Temporal                  & 12        & IFGoperc.R  & Frontal                   \\ \bottomrule
\end{tabular}
}
\label{tab:site_variant_con}
\end{table*}

\end{appendix}

\end{document}